# GIB: Imperfect Information in a Computationally Challenging Game

**Matthew L. Ginsberg**                                    GINSBERG@CIRL.UOREGON.EDU
*CIRL*
*1269 University of Oregon*
*Eugene, OR 97405 USA*

## Abstract

This paper investigates the problems arising in the construction of a program to play the game of contract bridge. These problems include both the difficulty of solving the game's perfect information variant, and techniques needed to address the fact that bridge is not, in fact, a perfect information game. GIB, the program being described, involves five separate technical advances: partition search, the practical application of Monte Carlo techniques to realistic problems, a focus on achievable sets to solve problems inherent in the Monte Carlo approach, an extension of alpha-beta pruning from total orders to arbitrary distributive lattices, and the use of squeaky wheel optimization to find approximately optimal solutions to cardplay problems.

GIB is currently believed to be of approximately expert caliber, and is currently the strongest computer bridge program in the world.

## 1. Introduction

Of all the classic games of mental skill, only card games and Go have yet to see the appearance of serious computer challengers. In Go, this appears to be because the game is fundamentally one of pattern recognition as opposed to search; the brute-force techniques that have been so successful in the development of chess-playing programs have failed almost utterly to deal with Go's huge branching factor. Indeed, the arguably strongest Go program in the world (Handtalk) was beaten by 1-dan Janice Kim (winner of the 1984 Fuji Women's Championship) in the 1997 AAAI Hall of Champions after Kim had given the program a monumental 25 stone handicap.

Card games appear to be different. Perhaps because they are games of imperfect information, or perhaps for other reasons, existing poker and bridge programs are extremely weak. World poker champion Howard Lederer (Texas Hold'em, 1996) has said that he would expect to beat any existing poker program after five minutes' play.[1] Perennial world bridge champion Bob Hamman, seven-time winner of the Bermuda Bowl, summarized the state of bridge programs in 1994 by saying that, "They would have to improve to be hopeless."[†]

In poker, there is reason for optimism: the GALA system (Koller & Pfeffer, 1995), if applicable, promises to produce a computer player of unprecedented strength by reducing the poker "problem" to a large linear optimization problem which is then solved to generate a strategy that is nearly optimal in a game-theoretic sense. Schaeffer, author of the world champion checkers program Chinook (Schaeffer, 1997), is also reporting significant success in the poker domain (Billings, Papp, Schaeffer, & Szafron, 1998).

---

1. Many of the citations here are the results of personal communications. Such communications are indicated simply by the presence of a [†] in the accompanying text.

The situation in bridge has been bleaker. In addition, because the American Contract Bridge League (ACBL) does not rank the bulk of its players in meaningful ways, it is difficult to compare the strengths of competing programs or players.

In general, performance at bridge is measured by playing the same deal twice or more, with the cards held by one pair of players being given to another pair during the replay and the results then being compared.[2] A "team" in a bridge match thus typically consists of two pairs, with one pair playing the North/South (N/S) cards at one table and the other pair playing the E/W cards at the other table. The results obtained by the two pairs are added; if the sum is positive, the team wins this particular deal and if negative, they lose it.

In general, the numeric sum of the results obtained by the two pairs is converted to International Match Points, or IMPs. The purpose of the conversion is to diminish the impact of single deals on the total, lest an abnormal result on one particular deal have an unduly large impact on the result of an entire match.

Jeff Goldsmith[†] reports that the standard deviation on a single deal in bridge is about 5.5 IMPs, so that if two roughly equal pairs were to play the deal, it would not be surprising if one team beat the other by about this amount. It also appears that the difference between an average club player and an expert is about 1.5 IMPs (per deal played); the strongest players in the world are approximately 0.5 IMPs/deal better still. Excepting GIB, the strongest bridge playing programs appear to be slightly weaker than average club players.

Progress in computer bridge has been slow. An incorporation of planning techniques into Bridge Baron, for example, appears to have led to a performance increment of approximately 1/3 IMP per deal (Smith, Nau, & Throop, 1996). This modest improvement still leaves Bridge Baron far shy of expert-level (or even good amateur-level) performance.

Prior to 1997, bridge programs generally attempted to duplicate human bridge-playing methodology in that they proceeded by attempting to recognize the class into which any particular deal fell: finesse, end play, squeeze, etc. Smith et al.'s work on the Bridge Baron program uses planning to extend this approach, but the plans continue to be constructed from human bridge techniques. Nygate and Sterling's early work on PYTHON (Sterling & Nygate, 1990) produced an expert system that could recognize squeezes but not prepare for them. In retrospect, perhaps we should have expected this approach to have limited success; certainly chess-playing programs that have attempted to mimic human methodology, such as PARADISE (Wilkins, 1980), have fared poorly.

GIB, introduced in 1998, works differently. Instead of modeling its play on techniques used by humans, GIB uses brute-force search to analyze the situation in which it finds itself. A variety of techniques are then used to suggest plays based on the results of the brute-force search. This technique has been so successful that all competitive bridge programs have switched from a knowledge-based approach to a search-based approach.

GIB's cardplay based on brute-force techniques was at the expert level (see Section 3) even without some of the extensions that we discuss in Section 5 and subsequently. The weakest part of GIB's game is bidding, where it relies on a large database of rules describing the meanings of various auctions. Quantitative comparisons here are difficult, although the general impression of the stronger players using GIB are that its overall play is comparable to that of a human expert.

---

2. The rules of bridge are summarized in Appendix A.

This paper describes the various techniques that have been used in the GIB project, as follows:

1. GIB's analysis in both bidding and cardplay rests on an ability to analyze bridge's perfect-information variant, where all of the cards are visible and each side attempts to take as many tricks as possible (this perfect-information variant is generally referred to as *double dummy* bridge). Double dummy problems are solved using a technique known as *partition search*, which is discussed in Section 2.

2. Early versions of GIB used *Monte Carlo methods* exclusively to select an action based on the double dummy analysis. This technique was originally proposed for cardplay by Levy (Levy, 1989), but was not implemented in a performance program before GIB. Extending Levy's suggestion, GIB uses Monte Carlo simulation for both cardplay (discussed in Section 3) and bidding (discussed in Section 4).

3. Section 5 discusses difficulties with the Monte Carlo approach. Frank et al. have suggested dealing with these problems by searching the space of possible plans for playing a particular bridge deal, but their methods appear to be intractable in both theory and practice (Frank & Basin, 1998; Frank, Basin, & Bundy, 2000). We instead choose to deal with the difficulties by modifying our understanding of the game so that the value of a bridge deal is not an integer (the number of tricks that can be taken) but is instead taken from a distributive lattice.

4. In Section 6, we show that the alpha-beta pruning mechanism can be extended to deal with games of this type. This allows us to find optimal plans for playing bridge end positions involving some 32 cards or fewer. (In contrast, Frank's method is capable only of finding solutions in 16 card endings.)

5. Finally, applying our ideas to the play of full deals (52 cards) requires solving an approximate version of the overall problem. In Section 7, we describe the nature of the approximation used and our application of *squeaky wheel optimization* (Joslin & Clements, 1999) to solve it.

Concluding remarks are contained in Section 8.

## 2. Partition search

Computers are effective game players only to the extent that brute-force search can overcome innate stupidity; most of their time spent searching is spent examining moves that a human player would discard as obviously without merit.

As an example, suppose that White has a forced win in a particular chess position, perhaps beginning with an attack on Black's queen. A human analyzing the position will see that if Black doesn't respond to the attack, he will lose his queen; the analysis considers places to which the queen could move and appropriate responses to each.

A machine considers responses to the queen moves as well, of course. But it must also analyze in detail every *other* Black move, carefully demonstrating that each of these other moves can be refuted by capturing the Black queen. A six-ply search will have to analyze every one of these moves five further ply, even if the refutations are identical in all cases. Conventional pruning techniques cannot help here; using $\alpha$-$\beta$ pruning, for example, the entire "main line" (White's winning choices and all of Black's losing responses) must be analyzed even though there is a great deal of apparent redundancy in this analysis.[3]

In other search problems, techniques based on the ideas of dependency maintenance (Stallman & Sussman, 1977) can potentially be used to overcome this sort of difficulty. As an example, consider chronological backtracking applied to a map coloring problem. When a dead end is reached and the search backs up, no information is cached and the effect is to eliminate only the specific dead end that was encountered. Recording information giving the reason for the failure can make the search substantially more efficient.

In attempting to color a map with only three colors, for example, thirty countries may have been colored while the detected contradiction involves only five. By recording the contradiction for those five countries, dead ends that fail for the same reason can be avoided.

Dependency-based methods have been of limited use in practice because of the overhead involved in constructing and using the collection of accumulated reasons. This problem has been substantially addressed in the work on dynamic backtracking (Ginsberg, 1993) and its successors such as RELSAT (Bayardo & Miranker, 1996), where polynomial limits are placed on the number of nogoods being maintained.

In game search, however, most algorithms already include significant cached information in the form of a transposition table (Greenblatt, Eastlake, & Crocker, 1967; Marsland, 1986). A transposition table stores a single game position and the backed up value that has been associated with it. The name reflects the fact that many games "transpose" in that identical positions can be reached by swapping the order in which moves are made. The transposition table eliminates the need to recompute values for positions that have already been analyzed.

These collected observations lead naturally to the idea that transposition tables should store not single positions and their values, but *sets* of positions and their values. Continuing the dependency-maintenance analogy, a transposition table storing sets of positions can prune the subsequent search far more efficiently than a table that stores only singletons.

There are two reasons that this approach works. The first, which we have already mentioned, is that most game-playing programs already maintain transposition tables, thereby incurring the bulk of the computational expense involved in storing such tables in a more general form. The second and more fundamental reason is that when a game ends with one player the winner, the reason for the victory is generally a local one. A chess game can be thought of as ending when one side has its king captured (a completely local phenomenon); a checkers game, when one side runs out of moves. Even if an internal search node is evaluated before the game ends, the reason for assigning it any specific value is likely to be independent of global features (e.g., is the Black pawn on $a5$ or $a6$?). Partition search exploits both the existence of transposition tables and the locality of evaluation for realistic games.

---

3. An informal solution to this is Adelson-Velskiy et al.'s *method of analogies* (Adelson-Velskiy, Arlazarov, & Donskoy, 1975). This approach appears to have been of little use in practice because it is restricted to a specific class of situations arising in chess games.

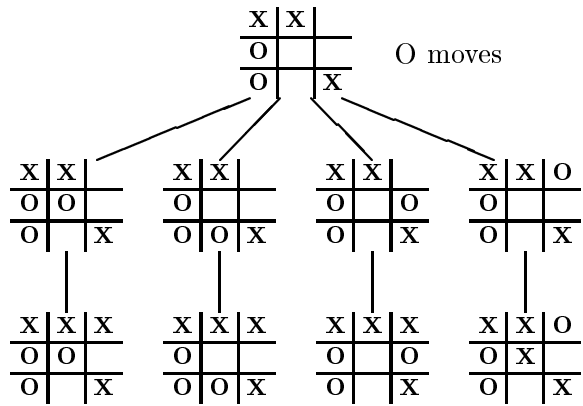

Figure 1: A portion of the game tree for tic-tac-toe

This section explains these ideas via an example and then describes them formally. Experimental results for bridge are also presented.

## 2.1 An example

Our illustrative examples for partition search will be taken from the game of tic-tac-toe. A portion of the game tree for this game appears in Figure 1, where we are analyzing a position that is a win for X. We show O's four possible moves, and a winning response for X in each case. Although X frequently wins by making a row across the top of the diagram, $\alpha$-$\beta$ pruning cannot reduce the size of this tree because O's losing options must all be analyzed separately.

Consider now the position at the lower left in the diagram, where X has won:

$$
\begin{array}{|c|c|c|}
\hline
\text{X} & \text{X} & \text{X} \\
\hline
\text{O} & \text{O} & \\
\hline
\text{O} & & \text{X} \\
\hline
\end{array}
\qquad (1)
$$

The reason that X has won is local. If we are retaining a list of positions with known outcomes, the entry we can make because of this position is:

$$
\begin{array}{|c|c|c|}
\hline
\text{X} & \text{X} & \text{X} \\
\hline
? & ? & ? \\
\hline
? & ? & ? \\
\hline
\end{array}
\qquad (2)
$$

where the ? means that it is irrelevant whether the associated square is marked with an X, an O, or unmarked. This table entry corresponds not to a single position, but to approximately $3^6$ because the unassigned squares can contain X's, O's, or be blank. We can reduce the game tree in Figure 1 to:

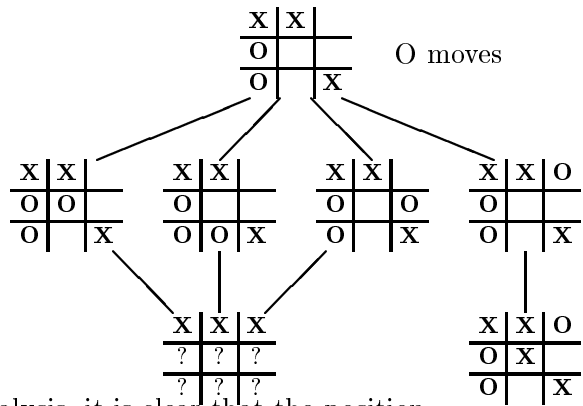

Continuing the analysis, it is clear that the position

$$\begin{array}{|c|c|c|} \hline \mathbf{X} & \mathbf{X} & \\ \hline ? & ? & ? \\ \hline ? & ? & ? \\ \hline \end{array} \qquad (3)$$

is a win for X if X is on play.[4] So is

$$\begin{array}{|c|c|c|} \hline \mathbf{X} & ? & ? \\ \hline ? & & ? \\ \hline ? & ? & \mathbf{X} \\ \hline \end{array}$$

and the tree can be reduced to:

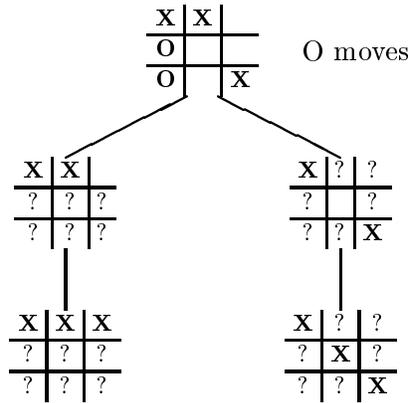

Finally, consider the position

$$\begin{array}{|c|c|c|} \hline \mathbf{X} & \mathbf{X} & \\ \hline ? & & \\ \hline ? & ? & \mathbf{X} \\ \hline \end{array} \qquad (4)$$

where it is O's turn as opposed to X's. If O moves in the second row, we get an instance of

$$\begin{array}{|c|c|c|} \hline \mathbf{X} & \mathbf{X} & \\ \hline ? & ? & ? \\ \hline ? & ? & ? \\ \hline \end{array}$$

while if O moves to the upper right, we get an instance of

$$\begin{array}{|c|c|c|} \hline \mathbf{X} & ? & ? \\ \hline ? & & ? \\ \hline ? & ? & \mathbf{X} \\ \hline \end{array}$$

---

4. We assume that O has not already won the game here, since X would not be "on play" if the game were over.

Thus every one of O's moves leads to a position that is known to be a win for X, and we can conclude that the original position (4) is a win for X as well. The root node in the reduced tree can therefore be replaced with the position of (4).

These positions capture the essence of the algorithm we will propose: If player $x$ can move to a position that is a member of a set known to be a win for $x$, the given position is a win as well. If every move is to a position that is a loss, the original position is also.

## 2.2 Formalizing partition search

In this section, we present a summary of existing methods for evaluating positions in game trees. There is nothing new here; our aim is simply to develop a precise framework in which our new results can be presented.

**Definition 2.2.1** *An* interval-valued game *is a quadruple* $(G, p_I, s, \texttt{ev})$*, where* $G$ *is a finite set of legal positions,* $p_I \in G$ *is the initial position,* $s : G \to 2^G$ *gives the immediate successors of a given position, and* $\texttt{ev}$ *is an evaluation function*

$$\texttt{ev} : G \to \{\texttt{max}, \texttt{min}\} \cup [0, 1]$$

*Informally,* $p' \in s(p)$ *means that position* $p'$ *can be reached from* $p$ *in a single move, and the evaluation function* $\texttt{ev}$ *labels internal nodes based upon whose turn it is to play (*$\texttt{max}$ *or* $\texttt{min}$*) and values terminal positions in terms of some element of the unit interval* $[0, 1]$*.*

*The structures* $G$*,* $p_I$*,* $s$ *and* $\texttt{ev}$ *are required to satisfy the following conditions:*

1. *There is no sequence of positions* $p_0, \ldots, p_n$ *with* $n > 0$*,* $p_i \in s(p_{i-1})$ *for each* $i$ *and* $p_n = p_0$*. In other words, there are no "loops" that return to an identical position.*

2. $\texttt{ev}(p) \in [0, 1]$ *if and only if* $s(p) = \emptyset$*. In other words,* $\texttt{ev}$ *assigns a numerical value to* $p$ *if and only if the game is over. Informally,* $\texttt{ev}(p) = \texttt{max}$ *means that the maximizer is to play and* $\texttt{ev}(p) = \texttt{min}$ *means that the minimizer is to play.*

We use $2^G$ to denote the power set of $G$, the set of subsets of $G$. There are two further things to note about this definition.

First, the requirement that the game have no "loops" is consistent with all modern games. In chess, for example, positions can repeat but there is a concealed counter that draws the game if either a single position repeats three times or a certain number of moves pass without a capture or a pawn move. In fact, dealing with the hidden counter is more natural in a partition search setting than a conventional one, since the evaluation function is in general (although not always) independent of the value of the counter.

Second, the range of $\texttt{ev}$ includes the entire unit interval $[0, 1]$. The value 0 represents a win for the minimizer, and 1 a win for the maximizer. The intermediate values might correspond to intermediate results (e.g., a draw) or, more importantly, allow us to deal with internal search nodes that are being treated as terminal and assigned approximate values because no time remains for additional search.

The evaluation function $\texttt{ev}$ can be used to assign numerical values to the entire set $G$ of positions:

**Definition 2.2.2** *Given an interval-valued game* $(G, p_I, s, \mathtt{ev})$, *we introduce a function* $\mathtt{ev}_c : G \to [0, 1]$ *defined recursively by*

$$\mathtt{ev}_c(p) = \begin{cases} \mathtt{ev}(p), & \text{if } \mathtt{ev}(p) \in [0, 1]; \\ \max_{p' \in s(p)} \mathtt{ev}_c(p'), & \text{if } \mathtt{ev}(p) = \max; \\ \min_{p' \in s(p)} \mathtt{ev}_c(p'), & \text{if } \mathtt{ev}(p) = \min. \end{cases}$$

*The* value *of* $(G, p_I, s, \mathtt{ev})$ *is defined to be* $\mathtt{ev}_c(p_I)$.

To evaluate a position in a game, we can use the well-known *minimax* procedure:

**Algorithm 2.2.3 (Minimax)** For a game $(G, p_I, s, \mathtt{ev})$ and a position $p \in G$, to compute $\mathtt{ev}_c(p)$:

**if** $\mathtt{ev}(p) \in [0, 1]$ **return** $\mathtt{ev}(p)$
**if** $\mathtt{ev}(p) = \max$ **return** $\max_{p' \in s(p)} \mathtt{minimax}(p')$
**if** $\mathtt{ev}(p) = \min$ **return** $\min_{p' \in s(p)} \mathtt{minimax}(p')$

There are two ways in which the above algorithm is typically extended. The first involves the introduction of transposition tables; we will assume that a new entry is added to the transposition table $T$ whenever one is computed. (A modification to cache only selected results is straightforward.) The second involves the introduction of $\alpha$-$\beta$ pruning. Incorporating these ideas gives us the algorithm at the top of the next page.

Each entry in the transposition table consists of a position $p$, the current cutoffs $[x, y]$, and the computed value $v$. Note the need to include information about the cutoffs in the transposition table itself, since the validity of any particular entry depends on the cutoffs in question.

As an example, suppose that the value of some node is in fact 1 (a win for the maximizer) but that when the node is evaluated with cutoffs of $[0, 0.5]$ a value of 0.5 is returned (indicating a draw) because the maximizer has an obviously drawing line. It is clear that this value is only accurate for the given cutoffs; wider cutoffs will lead to a different answer.

In general, the upper cutoff $y$ is the currently smallest value assigned to a minimizing node; the minimizer can do at least this well in that he can force a value of $y$ or lower. Similarly, $x$ is the currently greatest value assigned to a maximizing node. These cutoff values are updated as the algorithm is invoked recursively in the lines responsible for setting $v_{\text{new}}$, the value assigned to a child of the current position $p$.

**Proposition 2.2.4** *Suppose that* $v = \alpha\beta(p, [x, y])$ *for each entry* $(p, [x, y], v)$ *in* $T$. *Then if* $\mathtt{ev}_c(p) \in [x, y]$, *the value returned by Algorithm 2.2.5 is* $\mathtt{ev}_c(p)$. $\quad\square$

**Algorithm 2.2.5 ($\alpha$-$\beta$ pruning with transposition tables)** Given an interval-valued game $(G, p_I, s, \mathtt{ev})$, a position $p \in G$, cutoffs $[x, y] \subseteq [0, 1]$ and a transposition table $T$ consisting of triples $(p, [a, b], v)$ with $p \in G$ and $a \le b, v \in [0, 1]$, to compute $\alpha\beta(p, [x, y])$:

**if** there is an entry $(p, [x, y], z)$ in $T$ **return** $z$
**if** $\mathtt{ev}(p) \in [0, 1]$ **then** $v_{\mathrm{ans}} = \mathtt{ev}(p)$
**if** $\mathtt{ev}(p) = \mathtt{max}$ **then**
 $v_{\mathrm{ans}} := 0$
 **for** each $p' \in s(p)$ **do**
  $v_{\mathrm{new}} = \alpha\beta(p', [\max(v_{\mathrm{ans}}, x), y])$
  **if** $v_{\mathrm{new}} \ge y$ **then**
   $T := T \cup (p, [x, y], v_{\mathrm{new}})$
   **return** $v_{\mathrm{new}}$
  **if** $v_{\mathrm{new}} > v_{\mathrm{ans}}$ **then** $v_{\mathrm{ans}} = v_{\mathrm{new}}$
**if** $\mathtt{ev}(p) = \mathtt{min}$ **then**
 $v_{\mathrm{ans}} := 1$
 **for** each $p' \in s(p)$ **do**
  $v_{\mathrm{new}} = \alpha\beta(p', [x, \min(v_{\mathrm{ans}}, y)])$
  **if** $v_{\mathrm{new}} \le x$ **then**
   $T := T \cup (p, [x, y], v_{\mathrm{new}})$
   **return** $v_{\mathrm{new}}$
  **if** $v_{\mathrm{new}} < v_{\mathrm{ans}}$ **then** $v_{\mathrm{ans}} = v_{\mathrm{new}}$
$T := T \cup (p, [x, y], v_{\mathrm{ans}})$
**return** $v_{\mathrm{ans}}$

## 2.3 Partitions

We are now in a position to present our new ideas. We begin by formalizing the idea of a position that can reach a known winning position or one that can reach only known losing ones.

**Definition 2.3.1** *Given an interval-valued game $(G, p_I, s, \mathtt{ev})$ and a set of positions $S \subseteq G$, we will say that the set of positions that can **reach** $S$ is the set of all $p$ for which $s(p) \cap S \ne \emptyset$. This set will be denoted $R_0(S)$. The set of positions **constrained** to reach $S$ is the set of all $p$ for which $s(p) \subseteq S$, and is denoted $C_0(S)$.*

These definitions should match our intuition; the set of positions that can reach a set $S$ is indeed the set of positions $p$ for which some element of $S$ is an immediate successor of $p$, so that $s(p) \cap S \ne \emptyset$. Similarly, a position $p$ is constrained to reach $S$ if *every* immediate successor of $p$ is in $S$, so that $s(p) \subseteq S$.

Unfortunately, it may not be feasible to construct the $R_0$ and $C_0$ operators explicitly; there may be no concise representation of the set of all positions that can reach $S$. In practice, this will be reflected in the fact that the data structures being used to describe the set $S$ may not conveniently describe the set $R_0(S)$ of all situations from which $S$ can be reached.

Now suppose that we are expanding the search tree itself, and we find ourselves analyzing a particular position $p$ that is determined to be a win for the maximizer because the maximizer can move from $p$ to the winning set $S$; in other words, $p$ is a win because it is in $R_0(S)$. We would like to record at this point that the set $R_0(S)$ is a win for the maximizer, but may not be able to construct or represent this set conveniently. We will therefore assume that we have some computationally effective way to approximate the $R_0$ and $C_0$ functions, in that we have (for example) a function $R$ that is a conservative implementation of $R_0$ in that if $R$ says we can reach $S$, then so we can:

$$R(p, S) \subseteq R_0(S)$$

$R(p, S)$ is intended to represent a set of positions that are "like $p$ in that they can reach the (winning) set $S$." Note the inclusion of $p$ as an argument to $R(p, S)$, since we certainly want $p \in R(p, S)$. We are about to cache the fact that every element of $R(p, S)$ is a win for the maximizer, and certainly want that information to include the fact that $p$ itself has been shown to be a win. Thus we require $p \in R(p, S)$ as well.

Finally, we need some way to generalize the information returned by the evaluation function; if the evaluation function itself identifies a position $p$ as a win for the maximizer, we want to have some way to generalize this to a wider set of positions that are also wins. We formalize this by assuming that we have some generalization function $P$ that "respects" the evaluation function in the sense that the value returned by $P$ is a set of positions that `ev` evaluates identically.

**Definition 2.3.2** *Let $(G, p_I, S, \mathtt{ev})$ be an interval-valued game. Let $f$ be any function with range $2^G$, so that $f$ selects a set of positions based on its arguments. We will say that $f$ respects the evaluation function $\mathtt{ev}$ if whenever $p, p' \in F$ for any $F$ in the range of $f$, $\mathtt{ev}(p) = \mathtt{ev}(p')$.*

*A partition system for the game is a triple $(P, R, C)$ of functions that respect $\mathtt{ev}$ such that:*

1. *$P : G \to 2^G$ maps positions into sets of positions such that for any position $p$, $p \in P(p)$.*

2. *$R : G \times 2^G \to 2^G$ accepts as arguments a position $p$ and a set of positions $S$. If $p \in R_0(S)$, so that $p$ can reach $S$, then $p \in R(p, S) \subseteq R_0(S)$.*

3. *$C : G \times 2^G \to 2^G$ accepts as arguments a position $p$ and a set of positions $S$. If $p \in C_0(S)$, so that $p$ is constrained to reach $S$, then $p \in C(p, S) \subseteq C_0(S)$.*

As mentioned above, the function $P$ tells us which positions are sufficiently "like" $p$ that they evaluate to the same value. In tic-tac-toe, for example, the position (1) where X has won with a row across the top might be generalized by $P$ to the set of positions

| X | X | X |
|---|---|---|
| ? | ? | ? |
| ? | ? | ? |

$$(5)$$

as in (2).

The functions $R$ and $C$ approximate $R_0$ and $C_0$. Once again turning to our tic-tac-toe example, suppose that we take $S$ to be the set of positions appearing in (5) and that $p$ is given by

|   |   |   |
|---|---|---|
| X | X |   |
| O | O |   |
| O |   | X |

so that $S$ can be reached from $p$. $R(p, S)$ might be

|   |   |   |
|---|---|---|
| X | X |   |
| ? | ? | ? |
| ? | ? | ? |

(6)

as in (3), although we could also take $R(p, S) = \{p\}$ or $R(p, S)$ to be

|   |   |   |
|---|---|---|
| X | X |   |
| O | O |   |
| O |   | X |

$\cup$

|   |   |   |
|---|---|---|
| X | X |   |
| ? | ? | ? |
| ? | ? | ? |

$\cup$

|   |   |   |
|---|---|---|
| X | X |   |
| ? | ? | ? |
| ? | ? | ? |

although this last union might be awkward to represent. Note again that $R$ and $C$ are functions of $p$ as well as $S$; the set returned must include the given position $p$ but can otherwise be expected to vary as $p$ does.

We will now modify Algorithm 2.2.5 so that the transposition table, instead of caching results for single positions, caches results for sets of positions. As discussed in the introduction to this section, this is an analog to the introduction of truth maintenance techniques into adversary search. The modified algorithm 2.3.3 appears in Figure 2 and returns a pair of values – the value for the given position, and a set of positions that will take the same value.

**Proposition 2.3.4** *Suppose that* $v = \alpha\beta(p, [x, y])$ *for every* $(S, [x, y], v)$ *in* $T$ *and* $p \in S$. *Then if* $\mathtt{ev}_c(p) \in [x, y]$, *the value returned by Algorithm 2.3.3 is* $\mathtt{ev}_c(p)$.

**Proof.** We need to show that when the algorithm returns, any position in $S_{\mathrm{ans}}$ will have the value $v_{\mathrm{ans}}$. This will ensure that the transposition table remains correct.

To see this, suppose that the node being expanded is a maximizing node; the minimizing case is dual. Suppose first that this node is a loss for the maximizer, having value 0.

In showing that the node is a loss, we will have examined successor nodes that are in sets denoted $S_{\mathrm{new}}$ in Algorithm 2.3.3; if the maximizer subsequently finds himself in a position from which he has no moves outside of the various $S_{\mathrm{new}}$, he will still be in a losing position. Since $S_{\mathrm{all}} = \cup S_{\mathrm{new}}$, the maximizer will lose in any position from which he is constrained to next move into an element of $S_{\mathrm{all}}$. Since every position in $C(p, S_{\mathrm{all}})$ has this property, it is safe to take $S_{\mathrm{ans}} = C(p, S_{\mathrm{all}})$. This is what is done in the first line with a dagger in the algorithm.

The more interesting case is where the eventual value of the node is nonzero; now in order for another node $n$ to demonstrably have the same value, the maximizer must have no new options at $n$, and must still have some move that achieves the value $v_{\mathrm{ans}}$ at $n$.

The first condition is identical to the earlier case where $v_{\mathrm{ans}} = 0$. For the second, note that any time the maximizer finds a new best move, we set $S_{\mathrm{ans}}$ to the set of positions that **Algorithm 2.3.3 (Partition search)** Given a game $(G, p_I, s, \mathtt{ev})$ and $(P, R, C)$ a partition system for it, a position $p \in G$, cutoffs $[x, y] \subseteq [0, 1]$ and a transposition table $T$ consisting of triples $(S, [a, b], v)$ with $S \subseteq G$ and $a \leq b, v \in [0, 1]$, to compute $\alpha\beta(p, [x, y])$:

**if** there is an entry $(S, [x, y], z)$ with $p \in S$ **return** $\langle z, S \rangle$
**if** $\mathtt{ev}(p) \in [0, 1]$ **then** $\langle v_{\text{ans}}, S_{\text{ans}} \rangle = \langle \mathtt{ev}(p), P(p) \rangle$
**if** $\mathtt{ev}(p) = \mathtt{max}$ **then**
    $v_{\text{ans}} := 0$
    $S_{\text{all}} := \varnothing$
    **for** each $p' \in s(p)$ **do**
        $\langle v_{\text{new}}, S_{\text{new}} \rangle = \alpha\beta(p', [\max(v_{\text{ans}}, x), y])$
        **if** $v_{\text{new}} \geq y$ **then**
            $T := T \cup (S_{\text{new}}, [x, y], v_{\text{new}})$
            **return** $\langle v_{\text{new}}, S_{\text{new}} \rangle$
        **if** $v_{\text{new}} > v_{\text{ans}}$ **then** $\langle v_{\text{ans}}, S_{\text{ans}} \rangle = \langle v_{\text{new}}, S_{\text{new}} \rangle$
        $S_{\text{all}} := S_{\text{all}} \cup S_{\text{new}}$
    **if** $v_{\text{ans}} = 0$ **then** $S_{\text{ans}} = C(p, S_{\text{all}})$         †
    **else** $S_{\text{ans}} = R(p, S_{\text{ans}}) \cap C(p, S_{\text{all}})$     †   ‡
**if** $\mathtt{ev}(p) = \mathtt{min}$ **then**
    $v_{\text{ans}} := 1$
    $S_{\text{all}} := \varnothing$
    **for** each $p' \in s(p)$ **do**
        $\langle v_{\text{new}}, S_{\text{new}} \rangle = \alpha\beta(p', [x, \min(v_{\text{ans}}, y)])$
        **if** $v_{\text{new}} \leq x$ **then**
            $T := T \cup (S_{\text{new}}, [x, y], v_{\text{new}})$
            **return** $\langle v_{\text{new}}, S_{\text{new}} \rangle$
        **if** $v_{\text{new}} < v_{\text{ans}}$ **then** $\langle v_{\text{ans}}, S_{\text{ans}} \rangle = \langle v_{\text{new}}, S_{\text{new}} \rangle$
        $S_{\text{all}} := S_{\text{all}} \cup S_{\text{new}}$
    **if** $v_{\text{ans}} = 1$ **then** $S_{\text{ans}} = C(p, S_{\text{all}})$
    **else** $S_{\text{ans}} = R(p, S_{\text{ans}}) \cap C(p, S_{\text{all}})$     ‡
$T := T \cup (S_{\text{ans}}, [x, y], v_{\text{ans}})$
**return** $\langle v_{\text{ans}}, S_{\text{ans}} \rangle$

Figure 2: The partition search algorithm

we know recursively achieve the same value. When we complete the maximizer's loop in the algorithm, it follows that $S_{\text{ans}}$ will be a set of positions from which the maximizer can indeed achieve the value $v_{\text{ans}}$. Thus the maximizer can also achieve that value from any position in $R(p, S_{\text{ans}})$. It follows that the overall set of positions known to have the value $v_{\text{ans}}$ is given by $R(p, S_{\text{ans}}) \cap C(p, S_{\text{all}})$, intersecting the two conditions of this paragraph. This is what is done in the second daggered step in the algorithm. □

## 2.4 Zero-window variations

The effectiveness of partition search depends crucially on the size of the sets maintained in the transposition table. If the sets are large, many positions will be evaluated by lookup. If the sets are small, partition search collapses to conventional $\alpha$-$\beta$ pruning.

An examination of Algorithm 2.3.3 suggests that the points in the algorithm at which the sets are reduced the most are those marked with a *double* dagger in the description, where an intersection is required because we need to ensure both that the player can make a move equivalent to his best one *and* that there are no other options. The effectiveness of the method would be improved if this possibility were removed.

To see how to do this, suppose for a moment that the evaluation function always returned 0 or 1, as opposed to intermediate values. Now if the maximizer is on play and the value $v_{\text{new}} = 1$, a prune will be generated because there can be no better value found for the maximizer. If all of the $v_{\text{new}}$ are 0, then $v_{\text{ans}} = 0$ and we can avoid the troublesome intersection. The maximizer loses and there is no "best" move that we have to worry about making.

In reality, the restriction to values of 0 or 1 is unrealistic. Some games, such as bridge, allow more than two outcomes, while others cannot be analyzed to termination and need to rely on evaluation functions that return approximate values for internal nodes. We can deal with these situations using a technique known as *zero-window search* (originally called *scout* search (Pearl, 1980)). To evaluate a specific position, one first estimates the value to be $e$ and then determines whether the actual value is above or below $e$ by treating any value $v > e$ as a win for the maximizer and any value $v \leq e$ as a win for the minimizer. The results of this calculation can then be used to refine the guess, and the process is repeated. If no initial estimate is available, a binary search can be used to find the value to within any desired tolerance.

Zero-window search is effective because little time is wasted on iterations where the estimate is wildly inaccurate; there will typically be many lines showing that a new estimate is needed. Most of the time is spent on the last iteration or two, developing tight bounds on the position being considered. There is an analog in conventional $\alpha$-$\beta$ pruning, where the bounds typically get tight quickly and the bulk of the analysis deals with a situation where the value of the original position is known to lie in a fairly narrow range.

In zero-window search, a node always evaluates to 0 or 1, since either $v > e$ or $v \leq e$. This allows a straightforward modification to Algorithm 2.3.3 that avoids the troublesome cases mentioned earlier.

## 2.5 Experimental results

Partition search was tested by analyzing 1000 randomly generated bridge deals and comparing the number of nodes expanded using partition search and conventional methods.

In addition to our general interest in bridge, there are two reasons why it can be expected that partition search will be useful for this game. First, partition search requires that the functions $R_0$ and $C_0$ support a partition-like analysis; it must be the case that an analysis of one situation will apply equally well to a variety of similar ones. Second, it must be possible to build approximating functions $R$ and $C$ that are reasonably accurate representatives of $R_0$ and $C_0$.

Bridge satisfies both of these properties. Expert discussion of a particular deal often will refer to small cards as $x$'s, indicating that it is indeed the case that the exact ranks of these cards are irrelevant. Second, it is possible to "back up" $x$'s from one position to its predecessors. If, for example, one player plays a club with no chance of having it impact the rest of the game, and by doing so reaches a position in which subsequent analysis shows him to have two small clubs, then he clearly must have had *three* small clubs originally. Finally, the fact that cards are simply being replaced by $x$'s means that it is possible to construct data structures for which the time per node expanded is virtually unchanged from that using conventional methods.

Perhaps an example will make this clearer. Consider the following partial bridge deal in which East is to lead and there are no trumps:

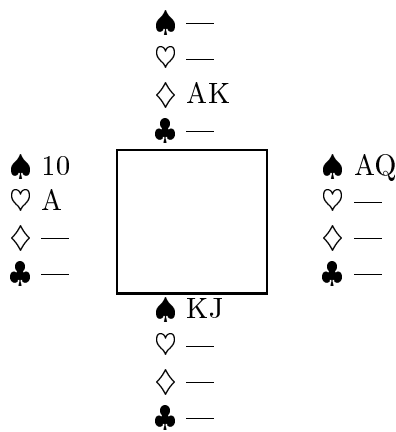

An analysis of this situation shows that in the main line, the only cards that win tricks by virtue of their ranks are the spade Ace, King and Queen. This sanctions the replacement of the above figure by the following more general one:

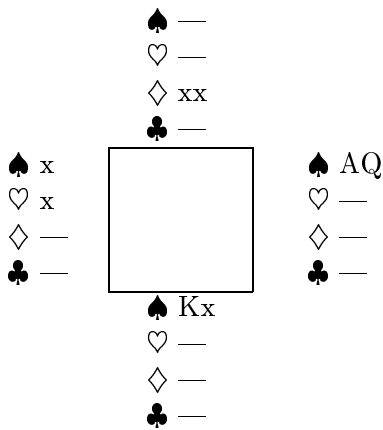

Note first that this replacement is *sound* in the sense that every position that is an instance of the second diagram is *guaranteed* to have the same value as the original. We have not resorted to an informal argument of the form "Jacks and lower tend not to matter," but instead to a precise argument of the form, "In the expansion of the search tree associated with the given deal, Jacks and lower were proven *never* to matter."

Bridge also appears to be extremely well-suited (no pun intended) to the kind of analysis that we have been describing; a chess analog might involve describing a mating combination and saying that "the position of Black's queen didn't matter." While this does happen, casual chess conversation is much less likely to include this sort of remark than bridge conversation is likely to refer to a host of small cards as $x$'s, suggesting at least that the partition technique is more easily applied to bridge than to chess (or to other games).

That said, however, the results for bridge are striking, leading to performance improvements of an order of magnitude or more on fairly small search spaces (perhaps $10^6$ nodes). The deals we tested involved between 12 and 48 cards and were analyzed to termination, so that the depth of the search varied from 12 to 48. (The solver without partition search was unable to solve larger problems.) The branching factor for minimax without transposition tables appeared to be approximately 4, and the results appear in Figure 3.

Each point in the graph corresponds to a single deal. The position of the point on the $x$-axis indicates the number of nodes expanded using $\alpha$-$\beta$ pruning and transposition tables, and the position on the $y$-axis the number expanded using partition search as well. Both axes are plotted logarithmically.

In both the partition and conventional cases, a binary zero-window search was used to determine the exact value to be assigned to the hand, which the rules of bridge constrain to range from 0 to the number of tricks left (one quarter of the number of cards in play). As mentioned previously, hands generated using a full deck of 52 cards were not considered because the conventional method was in general incapable of solving them. The program was run on a Sparc 5 and PowerMac 6100, where it expanded approximately 15K nodes/second. The transposition table shares common structure among different sets and as a result, uses approximately 6 bytes/node.

The dotted line in the figure is $y = x$ and corresponds to the breakeven point relative to $\alpha$-$\beta$ pruning in isolation. The solid line is the least-squares best fit to the logarithmic data, and is given by $y = 1.57x^{0.76}$. This suggests that partition search is leading to an effective reduction in branching factor of $b \to b^{0.76}$. This improvement, above and beyond that provided by $\alpha$-$\beta$ pruning, can be contrasted with $\alpha$-$\beta$ pruning itself, which gives a reduction when compared to pure minimax of $b \to b^{0.75}$ if the moves are ordered randomly (Pearl, 1982) and $b \to b^{0.5}$ if the ordering is optimal.

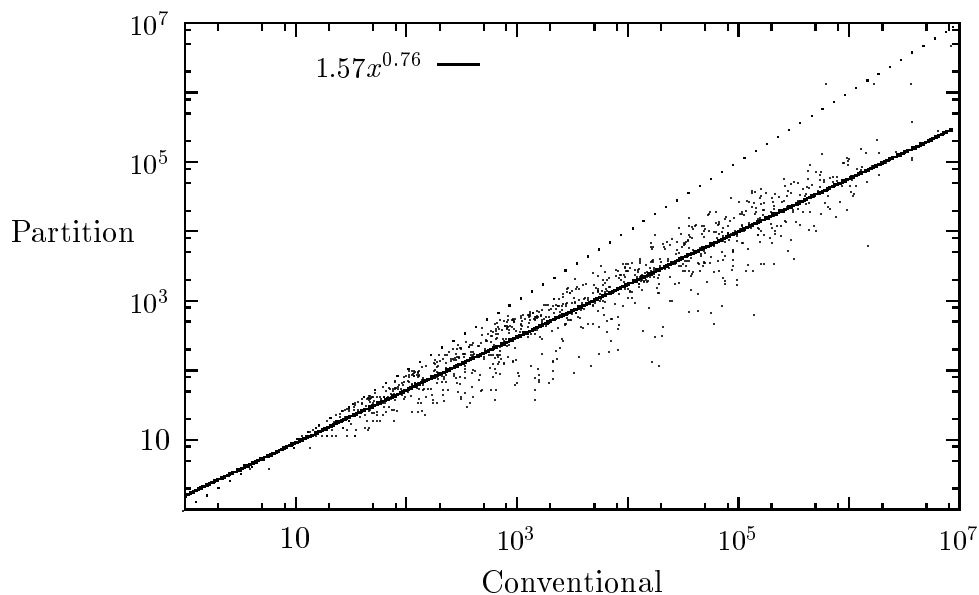

Figure 3: Nodes expanded as a function of method

The method was also applied to full deals of 52 cards, which can be solved while expanding an average of 18,000 nodes per deal.[5] This works out to about a second of CPU time.

## 3. Monte Carlo cardplay algorithms

One way in which we might use our perfect-information cardplay engine to proceed in a realistic situation would be to deal the unseen cards at random, biasing the deal so that it was consistent both with the bidding and with the cards played thus far. We could then analyze the resulting deal double dummy and decide which of our possible plays was the strongest. Averaging over a large number of such Monte Carlo samples would allow us to deal with the imperfect nature of bridge information. This idea was initially suggested by Levy (Levy, 1989), although he does not appear to have realized (see below) that there are problems with it in practice.

**Algorithm 3.0.1 (Monte Carlo card selection)** *To select a move from a candidate set M of such moves:*

---

5. The version of GIB that was released in October of 2000 replaced the transposition table with a data structure that uses a fixed amount of memory, and also sorts the moves based on narrowness (suggested by Plaat et al. (Plaat, Schaeffer, Pijls, & de Bruin, 1996) to be rooted in the idea of conspiracy search (McAllester, 1988)) and the killer heuristic. While the memory requirements are reduced, the overall performance is little changed.

1. *Construct a set $D$ of deals consistent with both the bidding and play of the deal thus far.*

2. *For each move $m \in M$ and each deal $d \in D$, evaluate the double dummy result of making the move $m$ in the deal $d$. Denote the score obtained by making this move $s(m, d)$.*

3. *Return that $m$ for which $\sum_d s(m, d)$ is maximal.*

The Monte Carlo approach has drawbacks that have been pointed out by a variety of authors, including Koller[†] and others (Frank & Basin, 1998). Most obvious among these is that the approach never suggests making an "information gathering play." After all, the perfect-information variant on which the decision is based invariably assumes that the information will be available by the time the next decision must be made! Instead, the tendency is for the approach to simply defer important decisions; in many situations this may lead to information gathering inadvertently, but the amount of information acquired will generally be far less than other approaches might provide.

As an example, suppose that on a particular deal, GIB has four possible lines of play to make its contract:

1. Line $A$ works if West has the ♠Q.

2. Line $B$ works if East has the ♠Q.

3. Line $C$ defers the guess until later.

4. Line $D$ (the clever line) works independent of who has the ♠Q.

Assuming that either player is equally likely to hold the ♠Q, a Monte Carlo analyzer will correctly conclude that line $A$ works half the time, and line $B$ works half the time. Line $C$, however, will be presumed to work *all* of the time, since the contract can still be made (double dummy) if the guess is deferred. Line $D$ will also be concluded to work all of the time (correctly, in this case).

As a result, GIB will choose randomly between the last two possibilities above, believing as it does that if it can only defer the guess until later (even the next card), it will make that guess correctly. The correct play, of course, is $D$.

We will discuss a solution to these difficulties in Sections 5–7; although GIB's defensive cardplay continues to be based on the above ideas, its declarer play now uses stronger techniques. Nevertheless, basing the card play on the algorithm presented leads to extremely strong results, approximately at the level of a human expert. Since GIB's introduction, all other competitive bridge-playing programs have switched their cardplay to similar methods, although GIB's double dummy analysis is substantially faster than most of the other programs and its play is correspondingly stronger.

We will describe three tests of GIB's cardplay algorithms: Performance on a commercially available set of benchmarks, performance in a human championship designed to highlight cardplay in isolation, and statistical performance measured over a large set of deals.

For the first test, we evaluated the strength of GIB's cardplay using *Bridge Master* (BM), a commercial program developed by Canadian internationalist Fred Gitelman. BM contains 180 deals at 5 levels of difficulty. Each of the 36 deals on each level is a problem in declarer play. If you misplay the hand, BM moves the defenders' cards around if necessary to ensure your defeat.

BM was used for the test instead of randomly dealt deals because the signal to noise ratio is far higher; good plays are generally rewarded and bad ones punished. Every deal also contains a lesson of some kind; there are no completely uninteresting deals where the line of play is irrelevant or obvious. There are drawbacks to testing GIB's performance on non-randomly dealt deals, of course, since the BM deals may in some way not be representative of the problems a bridge player would actually encounter at the table.

The test was run under Microsoft Windows on a 200 MHz Pentium Pro. As a benchmark, Bridge Baron (BB) version 6 was also tested on the same deals using the same hardware.[6] BB was given 10 seconds to select each play, and GIB was given 90 seconds to play the entire deal with a maximum Monte Carlo sample size of 50.[7] New deals were generated each time a play decision needed to be made.

These numbers approximately equalized the computational resources used by the two programs; BB could in theory take 260 seconds per deal (ten seconds on each of 26 plays), but in practice took substantially less. GIB was given the auctions as well; there was no facility for doing this in BB. This information was critical on a small number of deals.

Here is how the two systems performed:

| Level | BB | GIB |
|---|---|---|
| 1 | 16 | 31 |
| 2 | 8 | 23 |
| 3 | 2 | 12 |
| 4 | 1 | 21 |
| 5 | 4 | 13 |
| Total | 33 | 100 |
| | 18.3% | 55.6% |

Each entry is the number of deals that were played successfully by the program in question.

GIB's mistakes are illuminating. While some of them involve failing to gather information, most are problems in combining multiple chances (as in case $D$ above). As BM's deals get more difficult, they more often involve combining a variety of possibly winning options and that is why GIB's performance falls off at levels 2 and 3.

At still higher levels, however, BM typically involves the successful development of complex end positions, and GIB's performance rebounds. This appeared to happen to BB as well, although to a much lesser extent. It was gratifying to see GIB discover for itself the complex end positions around which the BM deals are designed, and more gratifying still to witness GIB's discovery of a maneuver that had hitherto not been identified in the bridge literature, as described in Appendix B.

---

6. The current version is Bridge Baron 10 and could be expected to perform guardedly better in a test such as this. Bridge Baron 6 does not include the Smith enhancements (Smith et al., 1996).

7. GIB's Monte Carlo sample size is fixed at 50 in most cases, which provides a good compromise between speed of play and accuracy of result.

Experiments such as this one are tedious, because there is no text interface to a commercial program such as Bridge Master or Bridge Baron. As a result, information regarding the sensitivity of GIB's performance to various parameters tends to be only anecdotal.

GIB solves an additional 16 problems (bringing its total to 64.4%) given additional resources in the form of extra time (up to 100 seconds per play, although that time was very rarely taken), a larger Monte Carlo sample (100 deals instead of 50) and hand-generated explanations of the opponents' bids and opening leads. Each of the three factors appeared to contribute equally to the improved performance.

Other authors are reporting comparable levels of performance for GIB. Forrester, working with a different but similar benchmark (Blackwood, 1979), reports[8] that GIB solves 68% of the problems given 20 seconds/play, and 74% of them given 30 seconds/play. Deals where GIB has outplayed human experts are the topic of a series of articles in the Dutch bridge magazine *IMP* (Eskes, 1997, and sequels).[9] Based on these results, GIB was invited to participate in an invitational event at the 1998 world bridge championships in France; the event involved deals similar to Bridge Master's but substantially more difficult. GIB joined a field of 34 of the best card players in the world, each player facing twelve such problems over the course of two days. GIB was leading at the halfway mark, but played poorly on the second day (perhaps the pressure was too much for it), and finished twelfth.

The human participants were given 90 minutes to play each deal, although they were penalized slightly for playing slowly. GIB played each deal in about ten minutes, using a Monte Carlo sample size of 500; tests before the event indicated little or no improvement if GIB were allotted more time. Michael Rosenberg, the eventual winner of the contest and the pre-tournament favorite, in fact made one more mistake than did Bart Bramley, the second place finisher. Rosenberg played just quickly enough that Bramley's accumulated time penalties gave Rosenberg the victory. The scoring method thus favors GIB slightly.

Finally, GIB's performance was evaluated directly using records from actual play. These records are available from high levels of human competition (world and national championships, typically), so that it is possible to determine exactly how frequently humans make mistakes at the bridge table. In Figure 4, we show the frequency with which this data indicates that a human declarer, leading to the *n*th trick of a deal, makes a mistake that causes his contract to become unmakeable on a double-dummy basis. The *y* axis gives the frequency of the mistakes and is plotted logarithmically; as one would expect, play becomes more accurate later in the deal.

We also give similar data for GIB, based on large sample of deals that GIB played against itself. The error profiles of the two are quite similar.

Before turning to defensive play, let me point out that this method of analysis favors GIB slightly. Failing to make an information gathering play gets reflected in the above figure, since the lack of information will cause GIB to make a double-dummy mistake subsequently. But human declarers often work to give the defenders problems that exploit their relative lack of information, and that tactic is not rewarded in the above analysis. Similar results for defensive play appear in Figure 5.

---

8. Posting to rec.games.bridge on 14 July 1997.

9. http://www.imp-bridge.nl

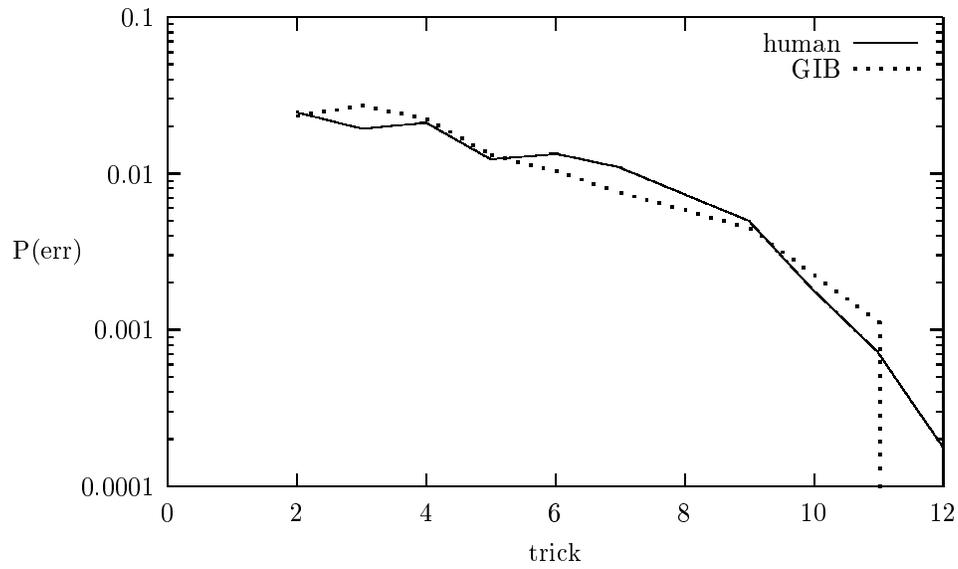

Figure 4: GIB's performance as declarer

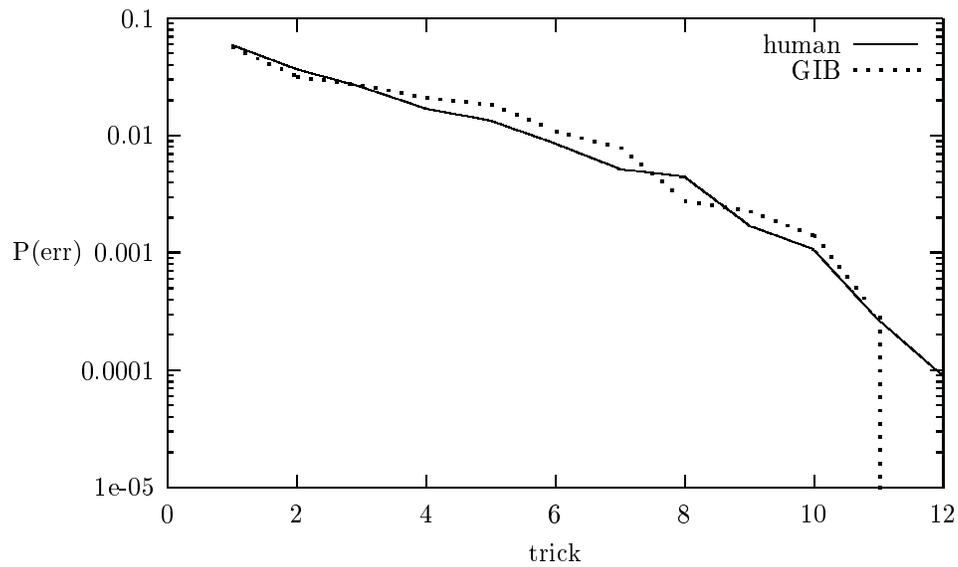

Figure 5: GIB's performance as defender

There are two important technical remarks that must be made about the Monte Carlo algorithm before proceeding. First, note that we were cavalier in simply saying, "Construct a set $D$ of deals consistent with both the bidding and play of the deal thus far."

To construct deals consistent with the bidding, we first simplify the auction as observed, building constraints describing each of the hands around the table. We then deal hands consistent with the constraints using a deal generator that deals unbiased hands given restrictions on the number of cards held by each player in each suit. This set of deals is then tested to remove elements that do not satisfy the remaining constraints, and each of the remaining deals is passed to the bidding module to identify those for which the observed bids would have been made by the players in question. (This assumes that GIB has a reasonable understanding of the bidding methods used by the opponents.) The overall dealing process typically takes one or two seconds to generate the full set of deals needed by the algorithm.

Now the card play must be analyzed. Ideally, GIB would do something similar to what it does for the bidding, determining whether each player would have played as indicated on any particular deal. Unfortunately, it is simply impractical to test each hypothetical decision recursively against the cardplay module itself. Instead, GIB tries to evaluate the probability that West (for example) has the ♠K (for example), and to then use these probabilities to weight the sample itself.

To understand the source of the weighting probabilities, let us consider a specific example. Suppose that in some particular situation, GIB plays the ♠5. The analysis indicates that 80% of the time that the next player (say West) holds the ♠K, it is a mistake for West not to play it. In other words, West's failure to play the ♠K leads to odds of 4:1 that he hasn't got it.

These odds are now used via Bayes' rule to adjust the probability that West holds the ♠K at all. The probabilities are then modified further to include information revealed by defensive signalling (if any), and the adjusted probabilities are finally used to bias the Monte Carlo sample. The evaluation $\sum_d s(m, d)$ in Algorithm 3.0.1 is replaced with $\sum_d w_d s(m, d)$ where $w_d$ is the weight assigned to deal $d$. More heavily weighted deals thus have a larger impact on GIB's eventual decision.

The second technical point regarding the algorithm itself involves the fact that it needs to run quickly and that it may need to be terminated before the analysis is complete. For the former, there are a variety of greedy techniques that can be used to ensure that a move $m$ is not considered if we can show $\sum_d s(d, m) \leq \sum_d s(d, m')$ for some $m'$. The algorithm also uses iterative broadening (Ginsberg & Harvey, 1992) to ensure that a low-width answer is available if a high-width search fails to terminate in time. Results from the low- and high-width searches are combined when time expires.

Also regarding speed, the algorithm requires that for each deal in the Monte Carlo sample and each possible move, we evaluate the resulting position exactly. Knowing simply that move $m_1$ is not as good as move $m_2$ for deal $d$ is not enough; $m_1$ may be better than $m_2$ elsewhere and we need to compare them quantitatively. This approach is aided substantially by the partition search idea, where entries in the transposition table correspond not to single positions and their evaluated values, but to sets of positions and values. In many cases, $m_1$ and $m_2$ may fall into the same entry of the partition table long before they actually transpose into one another exactly.

## 4. Monte Carlo bidding

The purpose of bidding in bridge is twofold. The primary purpose is to share information about your cards with your partner so that you can cooperatively select an optimal final contract. A secondary purpose is to disrupt the opponents' attempt to do the same.

In order to achieve this purpose, a wide variety of bidding "languages" have been developed. In some, when you suggest clubs as trumps, it means you have a lot of them. In others, the suggestion is only temporary and the information conveyed is quite different. In all of these languages, *some* meaning is assigned to a wide variety of bids in particular situations; there are also default rules that assign meanings to bids that have no specifically assigned meanings. Any computer bridge player will need similar understandings.

Bidding is interesting because the meanings frequently overlap; there may be one or more bids that are suitable (or nearly so) on any particular set of cards. Existing computer programs have simply matched possible bids against large databases giving their meanings, searching for that bid that best matches the cards that the machines hold. World champion Chip Martel reports[†] that human experts take a different approach.[10,11]

Although expert bidding is based on a database such as that used by existing programs, close decisions are made by simulating the results of each candidate action. This involves projecting how the bidding is likely to proceed and evaluating the play in one of a variety of possible final contracts. An expert gets his "judgment" from a Monte Carlo-like simulation of the results of possible bids, often referred to in the bridge-playing community as a *Borel* simulation (so named after the first player to describe the method). GIB takes a similar tack.

**Algorithm 4.0.2 (Borel simulation)** *To select a bid from a candidate set B, given a database Z that suggests bids in various situations:*

1. *Construct a set D of deals consistent with the bidding thus far.*

2. *For each bid $b \in B$ and each deal $d \in D$, use the database Z to project how the auction will continue if the bid b is made. (If no bid is suggested by the database, the player in question is assumed to pass.) Compute the double dummy result of the eventual contract, denoting it $s(b, d)$.*

3. *Return that b for which $\sum_d s(b, d)$ is maximal.*

As with the Monte Carlo approach to card play, this approach does not take into account the fact that bridge is not played double dummy. Human experts often choose not to make bids that will convey too much information to the opponents in order to make the defenders' task as difficult as possible. This consideration is missing from the above algorithm.[12]

---

10. The 1994 Rosenblum Cup World Team Championship was won by a team that included Martel and Rosenberg.

11. Frank suggests (Frank, 1998) that the existing machine approach is capable of reaching expert levels of performance. While this appears to have been true in the early 1980's (Lindelöf, 1983), modern expert bidding practice has begun to highlight the disruptive aspect of bidding, and machine performance is no longer likely to be competitive.

12. In theory at least, this issue could be addressed using the single-dummy ideas that we will present in subsequent sections. Computational considerations currently make this impractical, however.

There are more serious problems also, generally centering around the development of the bidding database $Z$.

First, the database itself needs to be built and debugged. A large number of rules need to be written, typically in a specialized language and dependent upon the bridge expertise of the author. The rules need to be debugged as actual play reveals oversights or other difficulties.

The nature and sizes of these databases vary enormously, although all of them represent very substantial investments on the part of the authors. The database distributed with MEADOWLARK BRIDGE includes some 7300 rules; that with Q-PLUS BRIDGE 2500 rules comprising 40,000 lines of specialized code. GIB's database is built using a derivative of the Meadowlark language, and includes about 3000 rules.

All of these databases doubtless contain errors of one sort or another; one of the nice things about most bidding methods is that they tend to be fairly robust against such problems. Unfortunately, the Borel algorithm described above introduces substantial instability in GIB's overall bidding.

To understand this, suppose that the database $Z$ is somewhat conservative in its actions. The projection in step 2 of Algorithm 4.0.2 now leads each player to assume its partner bids conservatively, and therefore to bid somewhat aggressively to compensate. The partnership as a whole ends up *over*compensating.

Worse still, suppose that there is an omission of some kind in $Z$; perhaps every time someone bids $7\diamondsuit$, the database suggests a foolish action. Since $7\diamondsuit$ is a rare bid, a bidding system that matches its bids directly to the database will encounter this problem infrequently.

GIB, however, will be much more aggressive, bidding $7\diamondsuit$ often on the grounds that doing so will cause the opponents to make a mistake. In practice, of course, the bug in the database is unlikely to be replicated in the opponents' minds, and GIB's attempts to exploit the gap will be unrewarded or worse.

This is a serious problem, and appears to apply to any attempt to heuristically model an adversary's behavior: It is difficult to distinguish a good choice that is successful because the opponent has no winning options from a bad choice that *appears* successful because the heuristic fails to identify such options.

There are a variety of ways in which this problem might be addressed, none of them perfect. The most obvious is simply to use GIB's aggressive tendencies to identify the bugs or gaps in the bidding database, and to fix them. Because of the size of the database, this is a slow process.

Another approach is to try to identify the bugs in the database automatically, and to be wary in such situations. If the bidding simulation indicates that the opponents are about to achieve a result much worse than what they might achieve if they saw each other's cards, that is evidence that there may be a gap in the database. Unfortunately, it is also evidence that GIB is simply effectively disrupting its opponents' efforts to bid accurately.

Finally, restrictions could be placed on GIB that require it to make bids that are "close" to the bids suggested by the database, on the grounds that such bids are more likely to reflect improvements in judgment than to highlight gaps in the database.

All of these techniques are used, and all of them are useful. GIB's bidding is substantially better than that of earlier programs, but not yet of expert caliber.

The bidding was tested as part of the 1998 Baron Barclay/OKBridge World Computer Bridge Championships, and the 2000 Orbis World Computer Bridge Championship. Each program bid deals that had previously been bid and played by experts; a result of 0 on any particular deal meant that the program bid to a contract as good as the average expert result. A positive result was better, and a negative result was worse.

There were 20 deals in each contest; although card play was not an issue, the deals were selected to pose challenges in bidding and a standard deviation of 5.5 IMPs/deal is still a reasonable estimate. One standard deviation over the 20 deal set could thus be expected to be about 25 IMPs.

GIB's final score in the 1998 bidding contest was +2 IMPs; in the 2000 contest it was +9 IMPs. In both cases, it narrowly edged out the expert field against which it was compared.[13] The next best program in 1998, Blue Chip Bridge, finished with a score of -35 IMPs, not dissimilar from the -37 IMPs that had been sufficient to win the bidding contest in 1997. The second place program in 2000 (once again Blue Chip Bridge) had a score of -2 IMPs.

## 5. The value of information

In previous sections of this paper, we have described Monte Carlo methods for dealing with the fact that bridge is a game of imperfect information, and have also described possible problems with this approach. We now turn to ways to overcomes some of these difficulties.

For the moment, let me assume that we replace bridge with a $\{0, 1\}$ game, so that we are interested only in the question of whether declarer makes his contract. Overtricks or extra undertricks are irrelevant. At least as a first approximation, bridge experts often look at hands this way, only subsequently refining the analysis.

If you ask such an expert why he took a particular line on a deal, he will often say something like, "I was playing for each opponent to have three hearts," or "I was playing for West to hold the spade queen." What he is reporting is that set of distributions of the unseen cards for which he was expecting to make the hand.

At some level, the expert is treating the value of the game not as zero or one (which it would be if he could see the unseen cards), but as a *function* from the set of possible distributions of unseen cards into $\{0, 1\}$. If we denote this set of distributions by $S$, the value of the game is thus a function

$$f : S \to \{0, 1\}$$

We will follow standard mathematical notation and denote the set $\{0, 1\}$ by 2 and denote the set of functions $f : S \to 2$ by $2^S$.

It is possible to extend max and min from the set $\{0, 1\}$ to $2^S$ in a pointwise fashion, so that, for example

$$\min(f, g)(s) = \min(f(s), g(s)) \tag{7}$$

for functions $f, g \in 2^S$ and a specific situation $s \in S$. The maximizing function is defined similarly.

---

13. This is in spite of the earlier remark that GIB's bidding is not of expert caliber. GIB was fortunate in the bidding contests in that most of the problems involved situations handled by the database. When faced with a situation that it does not understand, GIB's bidding deteriorates drastically.

As an example, suppose that in a particular situation, there is one line of play $f$ that wins if West has the ♠Q. There is another line of play $g$ that wins if East has exactly three hearts. Now $\min(f, g)$ is the line of play that wins just in case *both* West has the ♠Q and East has three hearts, while $\max(f, g)$ is the line of play that wins if either condition obtains.

It is important to realize that the set $2^S$ is not totally ordered by these max and min functions, like the unit interval is. Instead, $2^S$ is an instance of a mathematical structure known as a *lattice* (Grätzer, 1978, and Section 6). At this point, we note only that we can extend Definition 2.2.1 to any set with maximization and minimization operators:

**Definition 5.0.3** *A game is an octuple $(G, V, p_I, s, \text{ev}, f_+, f_-)$ such that:*

1. *$G$ is a finite set of possible positions in the game.*

2. *$V$ is the set of values for the game.*

3. *$p_I \in G$ is the initial position of the game.*

4. *$s : G \to 2^G$ gives the successors of a given position.*

5. *$\text{ev} : G \to \{\text{max}, \text{min}\} \cup V$ gives the value for terminal positions or indicates which player is to move for nonterminal positions.*

6. *$f_+ : \mathcal{P}(V) \to V$ and $f_- : \mathcal{P}(V) \to V$ are the combination functions for the maximizer and minimizer respectively.*

*The structures $G$, $V$, $p_I$, $s$ and $\text{ev}$ are required to satisfy the following conditions (unchanged from Definition 2.2.1):*

1. *There is no sequence of positions $p_0, \ldots, p_n$ with $n > 0$, $p_i \in s(p_{i-1})$ for each $i$ and $p_n = p_0$. In other words, there are no "loops" that return to an identical position.*

2. *$\text{ev}(p) \in V$ if and only if $s(p) = \varnothing$.*

This definition extends Definition 2.2.1 only in that the value set and combination functions have been generalized. A such, Definition 5.0.3 includes both "conventional" games in which the values are numeric and the combination functions are max/min, and our more general setting where the values are functional and the combination functions combine them as described above.

As usual, we can use the maximization and minimization functions to assign a value to the root of the tree:

**Definition 5.0.4** *Given a game $(G, V, p_I, s, \text{ev}, f_+, f_-)$, we introduce a function $\text{ev}_c : G \to V$ defined recursively by*

$$\text{ev}_c(p) = \begin{cases} \text{ev}(p), & \text{if } \text{ev}(p) \in V; \\ f_+\{\text{ev}_c(p')|p' \in s(p)\}, & \text{if } \text{ev}(p) = \text{max}; \\ f_-\{\text{ev}_c(p')|p' \in s(p)\}, & \text{if } \text{ev}(p) = \text{min}. \end{cases}$$

*The value of $(G, V, p_I, s, \text{ev}, f_+, f_-)$ is defined to be $\text{ev}_c(p_I)$.*

The definition is well founded because the game has no loops, and it is straightforward to extend the minimax algorithm 2.2.3 to this more general formalism. We will discuss extensions of $\alpha$-$\beta$ pruning in the next section.

To flesh out our previous informal description, we need to instantiate Definition 5.0.3. We do this by having the value of any particular node correspond to the set of positions where the maximizer can win:

1. The set $G$ of positions is a set of pairs $(p, Z)$ where $p$ is a position with only two of the four bridge hands visible (i.e., a position in the "single dummy" game), and $Z$ is that subset of $S$ (the set of situations) that is consistent both with $p$ and with the cards that were played to reach $p$ from the initial position.

2. The value set $V$ is $2^S$.

3. The initial position $p_I$ is $(p_0, S)$, where $p_0$ is the initial single-dummy position.

4. The successor function is described as follows:

   (a) If the declarer/maximizer is on play in the given position, the successors are obtained by enumerating the maximizer's legal plays and leaving the set $Z$ of situations unchanged.

   (b) If the minimizer is on play in the given position, the successors are obtained by playing any card $c$ that is legal in any element of $Z$ and then restricting $Z$ to that subset for which $c$ is in fact a legal play.

5. Terminal nodes are nodes where all cards have been played, and therefore correspond to single situations $s$, since the locations of all cards have been revealed. For such a terminal position, if the declarer has made his contract, the value is $S$ (the entire set of positions possible at the root). If the declarer has failed to make his contract, the value is $S - \{s\}$.

6. The maximization and minimization functions are computed pointwise, so that

$$f_+(U, V) = U \cup V$$

and

$$f_-(U, V) = U \cap V$$

Given an initial single-dummy situation $p$ corresponding to a set $S$ of situations, we will call the above game the $(p, S)$ *game*.

**Proposition 5.0.5** *Suppose that the set of situations for which the maximizer can make his contract is $T \subseteq S$. Then the value of the $(p, S)$ game is $T$.*

It is natural to view $T$ as an element of $2^S$; it is the function mapping points in $T$ to 1 and points outside of $T$ to 0.

**Proof.** The proof proceeds by induction on the depth of the game tree. If the root node $p$ is also terminal, then $S = \{s\}$ and the value is clearly set correctly (to $s$ or $\varnothing$) by the definition of the $(p, S)$ game.

If $p$ is nonterminal, suppose first that it is a maximizing node. Now let $s \in S$ be some particular situation. If the maximizer can win in $s$, then there is some successor $(p', S')$ to $(p, S)$ where the maximizer wins, and hence by the inductive hypothesis, the value of $(p', S')$ is a set $U$ with $s \in U$. But since the maximizer moves in $p$, the value assigned to $(p, S)$ is a superset of the value assigned to any subnode, so that $s \in \mathtt{ev}_c(p, S) = T$.

If, on the other hand, the maximizer cannot win in $s$, then he cannot win in any child of $s$. If $(p_i, S_i)$ are the successors of $(p, S)$ in the game tree, then again by the inductive hypothesis, we must have $s \notin \mathtt{ev}_c(p_i, S_i)$ for each $i$. But

$$\mathtt{ev}_c(p, S) = \cup_i \mathtt{ev}_c(p_i, S_i)$$

so that $s \notin \mathtt{ev}_c(p, S) = T$.

For the minimizing case, suppose that the maximizer wins in $s$. Then the maximizer must win in every successor of $s$, so that $s \in \mathtt{ev}_c(p_i, S_i)$ for each such successor and therefore $s \in \mathtt{ev}_c(p, S)$. Alternatively, if the minimizer wins in $s$, he must have a legal winning option so that $s \notin \mathtt{ev}_c(p_i, S_i)$ for some $i$ and therefore $s \notin \mathtt{ev}_c(p, S)$. □

Unfortunately, Proposition 5.0.5 is in some sense exactly what we wanted *not* to prove: it says that our modified game computes the set of situations in which it is possible for the maximizer to make his contract if he has perfect information about the opponents' cards, not the set of situations in which it is possible for him to make his contract given his actual state of incomplete information.

Before we go on to deal with this, however, let me look at an example in some detail. The example we will use is similar to that of Section 3 and involves a situation where the maximizer can make his contract if either West has the ♠Q or East has three hearts. I will denote by $S$ the set of situations where West has the ♠Q, and by $T$ the set where East has three hearts. It's possible to tie in the "defer the guess" example from Section 3 as well, so I will do that also. Here is the game tree for the game in question:

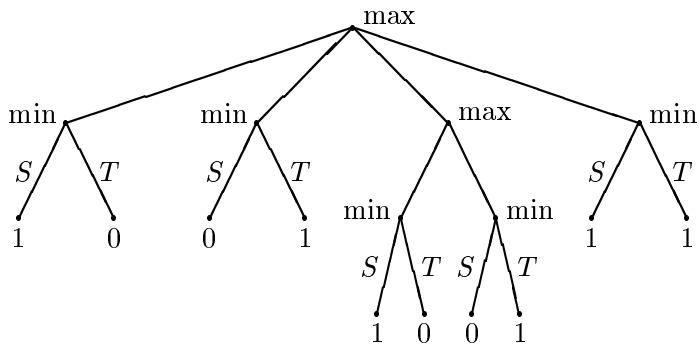

At the root node, the maximizer has four choices. If he makes the move on the left (playing for $S$, as it turns out), the minimizer then moves in a situation where the maximizer wins if $S$ holds and loses if $T$ holds. For the second move, where the maximizer is essentially playing for $T$, the reverse is true.

In the third case, the maximizer defers the guess. We suppose that he is on play again immediately, forced to commit between playing for $S$ and playing for $T$. In the last case, he wins independent of whether $T$ or $S$ obtains.

In the Monte Carlo setting, the above tree will actually be split based on the element of the sample in question. In some cases, $S$ will be true and we will examine only this subtree:

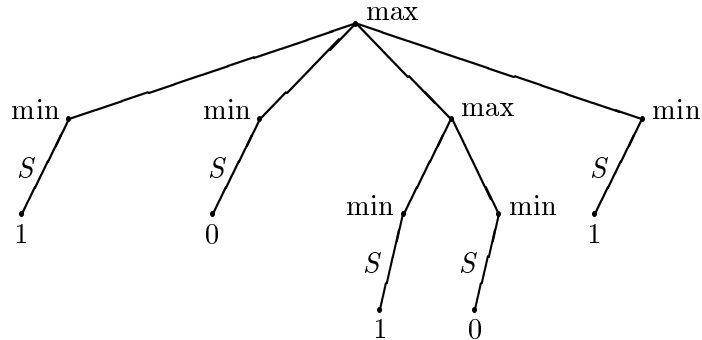

The maximizer can win by making any move other than the second. In the cases where $T$ obtains, we examine:

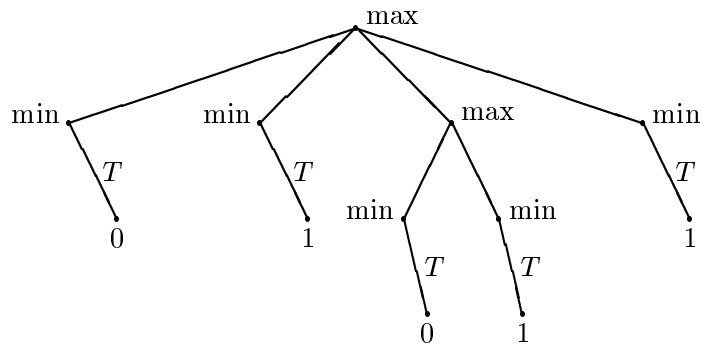

Here, the maximizer can win by making any move other than the first. In all cases, both of the last two moves win for the maximizer, since this approach cannot recognize the fact that the third move simply defers the guess while the fourth wins outright.

Now let us return to the situation where we include information about the sets that it is possible to play for. Here is the tree again:

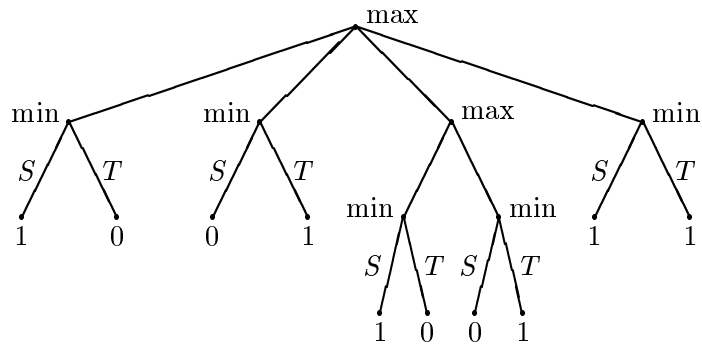

The first thing that we need to do is to realize that the terminal nodes should not be labelled with 1's and 0's but instead with sets where the maximizer can win. This produces:

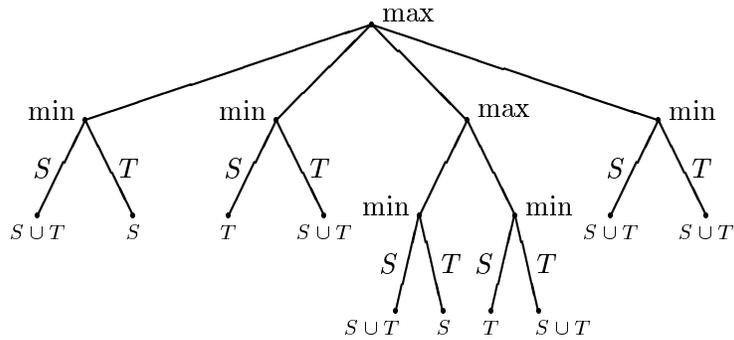

To understand the labels, consider the two leftmost fringe nodes. The leftmost node gets labelled with $T$ "for free" because $T$ is eliminated by the fact that the minimizer chose $S$. Since the maximizer wins in $S$, the maximizer wins in all cases.

For the second fringe node, $S$ is included by virtue of the minimizer's moving to $T$; $T$ is not included because the minimizer actually wins on this line. Hence the label of $T$ for the node in question. This analysis assumes that $S$ and $T$ are disjoint; if they overlap, the labels become slightly more complex but the overall analysis is little changed.

Backing up the values one step gives us:

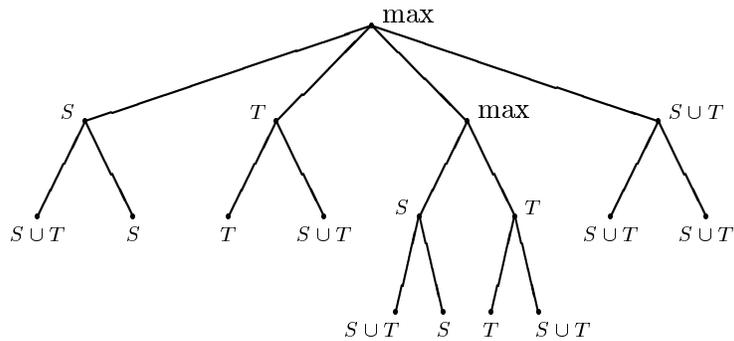

The minimizer, playing with perfect information, always does as best he can. The first interior node's label of $S$, for example, means that the maximizer wins only if $S$ actually is the case.

Of course, our definitions thus far imply that the maximizer is playing with perfect information as well, and we can back up the rest of the tree to get:

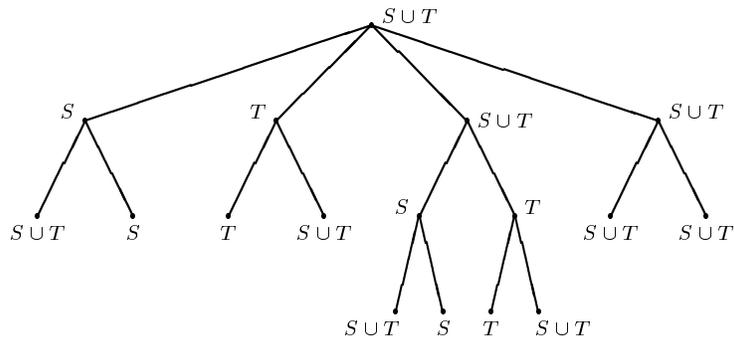

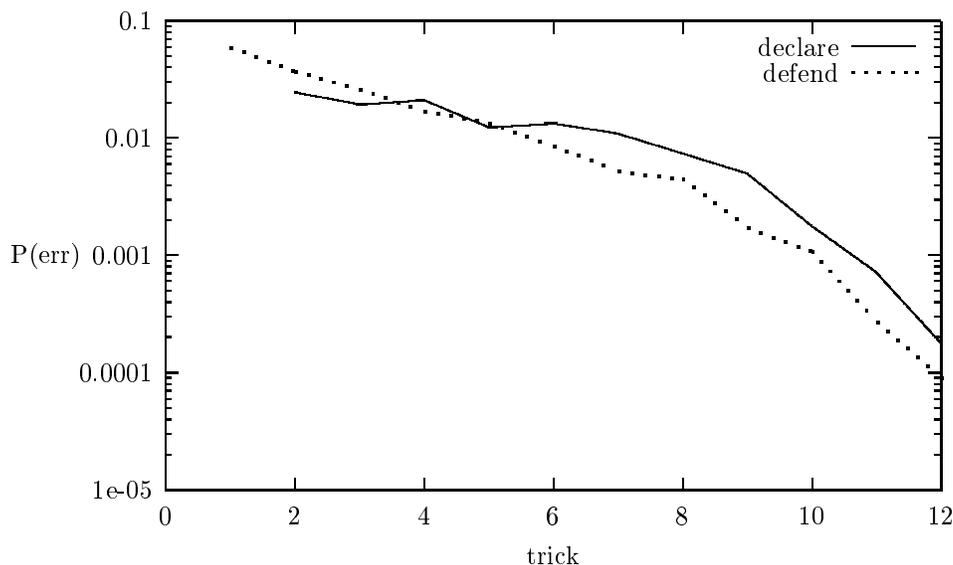

Figure 6: Defense vs. declarer play for humans

As before, the maximizer "wins" with either of the last two options.

Before we address the fact that the players do not in fact have perfect information, let me point out that in most bridge analyses, imperfect information is assumed to be an issue for the maximizer only. The defenders are assumed to be operating with complete information for at least the following reasons:

1. In general, there is a premium for declaring as opposed to defending, so that both sides want to declare. Typically, the pair with greater assets in terms of high cards wins the "bidding battle" and succeeds in becoming the declaring side, so that the overall assets available to the defenders in terms of high cards are generally less than those available to the declarer. This means that the defenders will generally be able to predict each other's hands with more accuracy than the declarer can.

2. The defenders can signal, conveying to one another information about the cards they hold. (As an example, play of an unnecessarily high card often indicates an even number of cards in the suit being played.) They are generally assumed to signal only information that is useful to them but not to declarer, once again improving their collective ability to play as if they had perfect information.

3. After the first two or three tricks, defenders' play is typically closer to double dummy than is the declarer's. This is shown in Figure 6, which contrasts the quality of human play as defender with the quality of human play as declarer; we make more mistakes declaring than defending as of trick four. (This figure is analogous to Figures 4 and 5.)

There are some deals where it is important for declarer to exploit uncertainty on the part of the defenders, but these are definitely the exception as opposed to the rule.

This suggests that Proposition 5.0.5 is doing a reasonable job of modeling the defenders' cardplay, but the combination function for the maximizer needs to be modified to reflect the imperfect-information nature of his task.

To understand this, let us return to our putative expert, who suggested at the beginning of this section that he might be playing for West to hold the spade queen. What he might say in a bit more detail is, "I could play for each opponent to hold exactly three hearts, or I could play for West to hold the spade queen. The latter was the better chance."

This suggests that the value assigned to the position by the maximizer is not a single set of situations (those in which he can make the contract), but a *set* $\mathcal{S}$ of sets of situations. Each set $S \in \mathcal{S}$ corresponds to one set of situations that the maximizer could play for, given his incomplete knowledge of the positions of the opposing cards.

Extending the notation used earlier in this section, we will denote the set of sets of situations by $2^{2^S}$. The maximizer's combination function on $2^{2^S}$ is given by

$$\max(\mathcal{F}, \mathcal{G}) = \mathcal{F} \cup \mathcal{G} \tag{8}$$

where each of $\mathcal{F}$ and $\mathcal{G}$ are sets of sets of situations. This says that if the maximizer is on play in a situation $p$, and he has one move that will allow him to select from a set $\mathcal{F}$ of things to "play for" and another move that will allow him to select from a set $\mathcal{G}$, then his choice at $p$ is to select from any element of $\mathcal{F} \cup \mathcal{G}$.

The minimizer's function is a bit more subtle. Suppose that at a node $p$, the minimizer can move to a successor with value $\mathcal{F} = \{F_i\}$, or to a successor with value $\mathcal{G} = \{G_i\}$. What value should we assign to $p$?

Since the minimizer has perfect information, he will always guarantee that the maximizer achieves the minimum value for the actual situation. Whatever element of $F_i \in \mathcal{F}$ or $G_j \in \mathcal{G}$ is eventually selected by the maximizer, the eventual value of $p$ will be the minimum of $F_i$ and $G_j$. In other words

$$\min(\{F_i\}, \{G_j\}) = \{\min(F_i, G_j)\} \tag{9}$$

where the individual minima are computed using the perfect information rule (7).

**Definition 5.0.6** *Let $G$ be the set of positions in an imperfect information game, a set of pairs $(p, Z)$ where $p$ is a position from the point of view of the maximizing player and $Z$ is the set of perfect information positions consistent with $p$. The* imperfect information game *for $G$ is the game $(G, V, p_I, s, \mathtt{ev}, f_+, f_-)$ where:*

1. *The value set $V$ is $2^{2^S}$.*

2. *The initial position $p_I$ is $(p_0, S)$, where $p_0$ is the initial imperfect information position and $S$ is the set of all perfect information positions consistent with it.*

3. *The successor function is described as follows:*

   (a) *If the maximizer is on play in the given position, the successors are obtained by enumerating the maximizer's legal plays and leaving the elements of the set $Z$ of situations unchanged.*

(b) *If the minimizer is on play in the given position, the successors are obtained by making playing any card $c$ that is legal in any element of $X$ and then restricting $Z$ to those situations for which $c$ is in fact a legal play.*

4. *Terminal nodes are nodes where all cards have been played, and therefore correspond to single situations $s$. For such a terminal position, if the declarer has made his contract, the value is $(\{s\}, \{S\})$. If the declarer has failed to make his contract, the value is $(\{s\}, \{S - \{s\}\})$.*

5. *The maximization and minimization functions are given by (8) and (9) respectively.*

**Theorem 5.0.7** *Suppose that the value of the imperfect information game for $G$ is $\mathcal{T}$. Then a set of positions $T$ is a subset of an element of $\mathcal{T}$ if and only if the maximizer has a strategy that wins in every element of $T$, assuming that the minimizer plays with perfect information.*

**Proof.** Once again, the proof proceeds by induction on the depth of the game tree. And once again, the case where $p$ is a terminal position is handled easily by the definition. For the inductive case, we consider the maximizer and minimizer separately.

For the maximizer, suppose that there is some set $T$ of situations that satisfies the conditions of the theorem, so that the maximizer has a strategy that caters to all of the elements of $T$. Then the first move of that strategy will be some single move to a position $p_i$ that is a successor of $p$ and that caters to the elements of $T$. Thus if the value of the successful child is $\mathcal{F}$, $T$ is a subset of some $F \in \mathcal{F}$ by the inductive hypothesis. Thus if the value of the original game is $\mathcal{G}$, $T$ is a subset of an element of $\mathcal{G}$ by virtue of (8).

Alternatively, if $T$ is a set for which the maximizer has no such strategy, then clearly the maximizer cannot have a strategy after making any of the moves to the successor positions $p_i$. This means that no superset $U \supseteq T$ in any $\mathtt{ev}_c(p_i)$, and thus no superset of $T$ in $\mathtt{ev}_c(p)$ either.

The minimizing case is not really any harder. Suppose first that the maximizer has no strategy for succeeding in every situation in $T$. Then the minimizer (playing with perfect information) must have some move to a position $p_i$ with value $\mathcal{F}_i$ such that $T$ is not a subset of any element of $\mathcal{F}_i$. Now if $\mathcal{F}_i = \{T_i\}$, recall that

$$\min(\{T_i\}, \{U_i\}) = \{T_i \cap U_j\},$$

and $T \not\subseteq T_i$ for each $i$. Thus $T \not\subseteq T_i \cap U_j$ for each $i$ and $j$, and there is no $V \supseteq T$ with

$$V \in \min(\{T_i\}, \{U_i\})$$

For the last case, suppose that the maximizer does have a strategy for succeeding in every situation in $T$. That means that after any move for the minimizer, the maximizer will still have a strategy that succeeds in $T$, so that if $p_i$ are the successors of $p$ and $\mathtt{ev}_c(p_i) = \mathcal{T}_i$, then there is a $T_i \in \mathcal{T}_i$ with $T \subseteq T_i$. Now $T \subseteq \cap_i T_i \in \min(\mathcal{T}_i) = \mathtt{ev}_c(p)$. Thus $\mathtt{ev}_c(p)$ contains an element that is a superset of $T$. $\square$

Using this result, we can in theory compute exactly the set of things we might play for given a single-dummy bridge problem. Before we turn to the issues involved in doing so in practice, however, let me repeat the example of this section using the imperfect information technique. Here is the game tree again with values assigned to the terminal nodes:

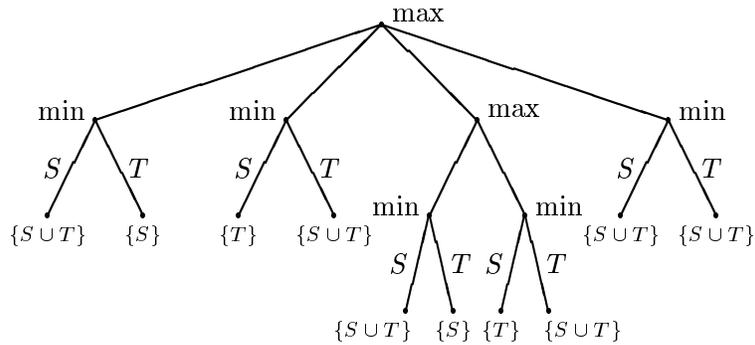

Backing up past the minimizer's final move gives us:

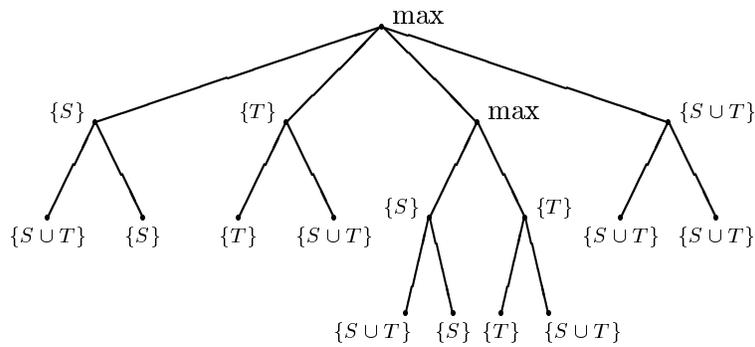

And we can now complete the analysis to finally get:

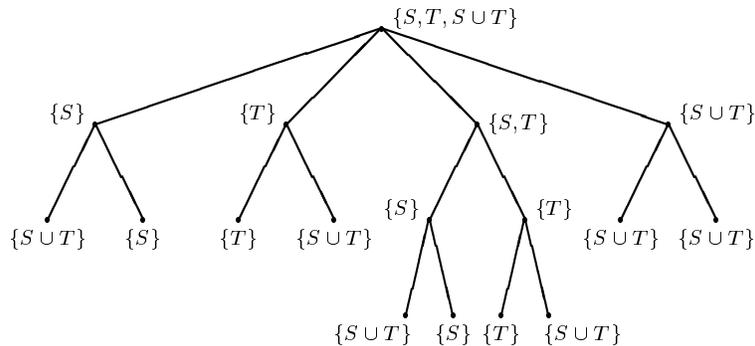

Note the difference in the values assigned to the maximizer's third and fourth choices at the first ply. The third choice has value $\{S, T\}$, indicating clearly that the maximizer will need to subsequently decide whether to play for $S$ or for $T$. But the fourth choice has value $\{S \cup T\}$ indicating that both possibilities are catered to.

The value assigned to the root contains some redundancy (which we will deal with in Section 7), in that one of the maximizer's choices ($S \cup T$) dominates the others. Nevertheless, this value clearly indicates that the maximizer has an option available at the root that caters to both situations.

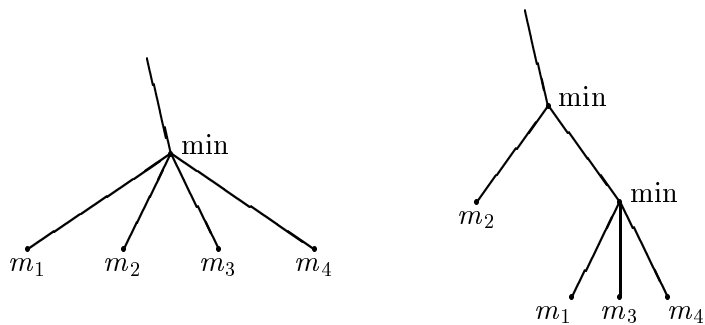

Figure 7: Equivalent games?

## 6. Extending alpha-beta pruning to lattices

The results of the previous section allow us to deal with imperfect information in theory. Unfortunately, computing the value in theory is hardly the same as computing it in practice. Some ideas, such as transposition tables and partition search, can fairly obviously be applied to games with values taken from sets more general than total orders. But what about $\alpha$-$\beta$ pruning, the linchpin of high-performance adversary search algorithms? The answer here is far more subtle.

### 6.1 Some necessary definitions

Let us begin by considering the two small game trees in Figure 7, where the minimizer is on play at the nonfringe nodes and none of the $m_i$ is intended to be necessarily terminal. Are these two games always equivalent?

We would argue that they are. In the game on the left, the minimizer needs to select among the four options $m_1, m_2, m_3, m_4$. In the game on the right, he needs to first select whether or not to play $m_2$; if he decides not to, he must select among the remaining options. Since the minimizer has the same possibilities in both cases, we assume that the values assigned to the games are the same.

From a more formal point of view, the value of the game on the left is $f_-(m_1, m_2, m_3, m_4)$, while that of the game on the right is $f_-(m_2, f_-(m_1, m_3, m_4))$ where we have abused notation somewhat, writing $m_i$ for the value of the node $m_i$ as well.

**Definition 6.1.1** *A game will be called* simple *if for any $x \in v \subseteq V$,*

$$f_+\{x\} = f_-\{x\} = x$$

*and also*

$$f_+(v) = f_+\{x, f_+(v - x)\}$$

*and*

$$f_-(v) = f_-\{x, f_-(v - x)\}$$

We have augmented the condition developed in the discussion of Figure 7 with the assumption that if a player's move in a position $p$ is forced (so that $p$ has a unique successor), then the value before and after the forced move is the same.

**Proposition 6.1.2** *For any simple game, there are binary functions $\wedge$ and $\vee$ from $V$ to itself that are commutative, associative and idempotent[14] and such that*

$$f_+\{v_0, \ldots, v_m\} = v_0 \vee \cdots \vee v_m$$

*and*

$$f_-\{v_0, \ldots, v_m\} = v_0 \wedge \cdots \wedge v_m.$$

**Proof.** Induction on $m$. □

When referring to a simple game, we will typically replace the functions $f_+$ and $f_-$ by the equivalent binary functions $\vee$ and $\wedge$. We assume throughout the rest of this section that all games are simple.[15]

The binary functions $\vee$ and $\wedge$ now induce a partial order $\leq$, where we will say that $x \leq y$ if and only if $x \vee y = y$. It is not hard to see that this partial order is reflexive ($x \leq x$), antisymmetric ($x \leq y$ and $y \leq x$ if and only if $x = y$) and transitive. The operators $\vee$ and $\wedge$ behave like greatest lower bound and least upper bound operators with regard to the partial order.

We also have the following:

**Proposition 6.1.3** *Whenever $S \subseteq T$, $f_+(S) \leq f_+(T)$ and $f_-(S) \geq f_-(T)$.* □

In other words, assuming that the minimizer is trying to reach a low value in the partial order and the maximizer is trying to reach a high one, having more options is always good.

## 6.2 Shallow pruning

We are now able to investigate $\alpha$-$\beta$ pruning in our general framework. Let us begin with shallow pruning, shown in Figure 8.

The idea here is that if the minimizer prefers $x$ to $y$, he will never allow the maximizer even the possibility of selecting between $y$ and the value of the subtree rooted at $T$. After all, the value of the maximizing node in the figure is $y \vee \mathtt{ev}_c(T) \geq y \geq x$, and the minimizer will therefore always prefer $x$.

In order for the usual correctness proof for (shallow) $\alpha$-$\beta$ pruning to hold, we need the following condition to be satisfied:

**Definition 6.2.1** *(Shallow $\alpha$-$\beta$ pruning) A game $G$ will be said to allow shallow $\alpha$-$\beta$ pruning for the minimizer if*

$$x \wedge (y \vee T) = x \tag{10}$$

---

14. A binary function $f$ is called *idempotent* if $f(a, a) = a$ for all a.

15. We also assume that the games are sufficiently complex that we can find in the game tree a node with any desired functional value, e.g., $a \wedge (b \vee c)$ for specific $a$, $b$ and $c$. Were this not the case, none of our results would follow. As an example, a game in which the initial position is also terminal surely admits pruning of all kinds (since the game tree is empty) but need not satisfy the conclusions of the results in this section.

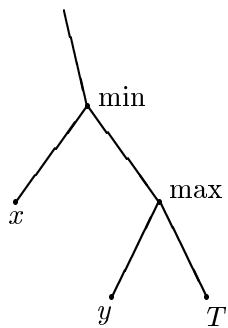

Figure 8: $T$ can be pruned (shallowly) if $x \leq y$

for all $x, y, T \in V$ with $x \leq y$. The game will be said to allow shallow $\alpha$-$\beta$ pruning for the maximizer if

$$x \vee (y \wedge T) = x \tag{11}$$

for all $x, y, T \in V$ with $x \geq y$. We will say that $G$ allows shallow pruning if it allows shallow $\alpha$-$\beta$ pruning for both players.

The definition basically says that the backed up value at the root of the game tree is unchanged by pruning the maximizing subtree in the figure.

As we will see shortly, the expressions (10) and (11) describing shallow pruning are identical to what are more typically known as *absorption identities*.

**Definition 6.2.2** *Suppose $V$ is a set and $\wedge$ and $\vee$ are two binary operators on $V$. The triple $(V, \wedge, \vee)$ is called a* lattice *if $\wedge$ and $\vee$ are idempotent, commutative and associative, and satisfy the* absorption identities *in that for any $x, y \in V$,*

$$x \vee (x \wedge y) = x \tag{12}$$
$$x \wedge (x \vee y) = x \tag{13}$$

We also have the following:

**Definition 6.2.3** *A lattice $(V, \wedge, \vee)$ is called* distributive *if $\wedge$ and $\vee$ distribute with respect to one another, so that*

$$x \vee (y \wedge z) = (x \vee y) \wedge (x \vee z) \tag{14}$$
$$x \wedge (y \vee z) = (x \wedge y) \vee (x \wedge z) \tag{15}$$

**Lemma 6.2.4** *Each of (12) and (13) implies the other. Each of (14) and (15) implies the other.*

**Proof.** These are well known results from lattice theory (Grätzer, 1978). □

**Proposition 6.2.5 (Ginsberg & Jaffray, 2001)** *For a game $G$, the following conditions are equivalent:*

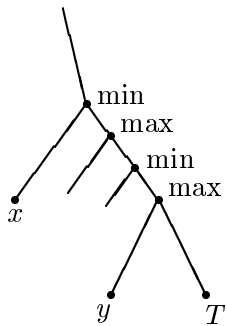

Figure 9: $T$ can be pruned (deeply) if $x \leq y$

1. *G allows shallow $\alpha$-$\beta$ pruning for the minimizer.*

2. *G allows shallow $\alpha$-$\beta$ pruning for the maximizer.*

3. *G allows shallow pruning.*

4. *$(V, \wedge, \vee)$ is a lattice.*

**Proof.**[16] We show that the first and fourth conditions are equivalent; everything else follows easily.

If $G$ allows shallow $\alpha$-$\beta$ pruning for the minimizer, we take $x = a$ and $y = T = a \vee b$ in (10). Clearly $x \leq y$ so we get

$$a \wedge (y \vee y) = a \wedge y = a \wedge (a \vee b) = a$$

as in (13).

For the converse, if $x \leq y$, then $x \wedge y = x$ and

$$
\begin{aligned}
x \wedge (y \vee T) & = (x \wedge y) \wedge (y \vee T) \\
& = x \wedge (y \wedge (y \vee T)) \\
& = x \wedge y \\
& = x. \quad \square
\end{aligned}
$$

### 6.3 Deep pruning

Deep pruning is a bit more subtle. An example appears in Figure 9.

As before, assume $x \leq y$. The argument is as described previously: Given that the minimizer has a guaranteed value of $x$ at the upper minimizing node, there is no way that a choice allowing the maximizer to reach $y$ can be on the main line; if it were, then the maximizer could get a value of at least $y$.

---

16. The proofs of this and Proposition 6.3.2 are due to Alan Jaffray.

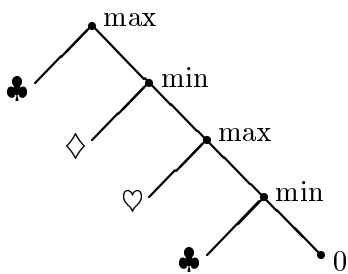

Figure 10: The deep pruning counterexample

**Definition 6.3.1** *(Deep $\alpha$-$\beta$ pruning) A game $G$ will be said to* allow $\alpha$-$\beta$ pruning for the minimizer *if for any $x, y, T, z_1, \ldots, z_{2i} \in V$ with $x \leq y$,*

$$x \wedge (z_1 \vee (z_2 \wedge \cdots \vee (z_{2i} \wedge (y \vee T))) \cdots) =$$
$$x \wedge (z_1 \vee (z_2 \wedge \cdots \vee z_{2i}) \cdots).$$

*The game will be said to* allow $\alpha$-$\beta$ pruning for the maximizer *if*

$$x \vee (z_1 \wedge (z_2 \vee \cdots \wedge (z_{2i} \vee (y \wedge T))) \cdots) =$$
$$x \vee (z_1 \wedge (z_2 \vee \cdots \wedge z_{2i}) \cdots).$$

*We will say that $G$* allows pruning *if it allows $\alpha$-$\beta$ pruning for both players.*

As before, the prune allows us to remove the dominated node ($y$ in Figure 9) and all of its siblings.

The fact that a game allows shallow $\alpha$-$\beta$ pruning does not mean that it allows pruning in general, as is shown by the following counterexample. The example involves a game with one card that is known to both players; only the suit of the card matters. The game tree appears in Figure 10.

In this tree, a node labelled with a suit symbol is terminal and means that the maximizer wins if and only if the suit of the card matches the given symbol. So at the root of the given tree, the maximizer (whose turn it is to play) can choose to "turn over" the card, winning if and only if it's a club, or can defer to the minimizer. The minimizer can choose to turn the card (*losing* just in case it's a diamond – the suit symbols refer to the maximizer's result), or hand the situation back to the maximizer. If the maximizer defers yet again, the minimizer can either turn over the card, losing if it's a club, or simply declare victory (presumably his choice).

There is one other wrinkle in this game. At any point in the game, the maximizer can change the card from either a diamond or a spade to a club.

Now let's consider the game itself. At ply 4, the minimizer will obviously choose to win the game. Thus at ply 3, the maximizer will need to choose $\heartsuit$, winning just in case the card is a heart. But this means that at ply 2, the minimizer will win the game, since if the card is not a diamond he will move to the left (and win at once) while if the card is not a heart he can win by moving to the right. (Remember that the minimizer knows the suit of the card.) The upshot of this is that the maximizer wins the overall game if and only if the card in question is a club. A formal analysis proceeds similarly, labelling the nodes as follows:

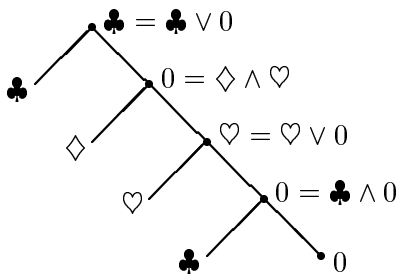

Note, incidentally, that the maximizer's ability to change the card does not help him win the game.

Now suppose that we apply deep pruning to this game. The ply four node is one where the minimizer can force a value of at most ♣, suggesting that the siblings of the bottom ♣ node can be pruned. But doing so produces the following tree:

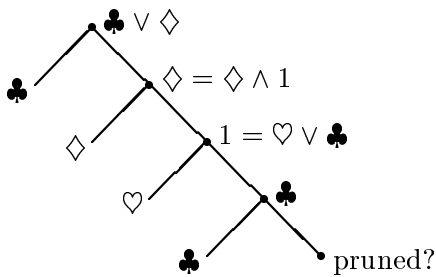

If the maximizer reaches ply 3, he can win by changing the card to a club if need be.

Of course, the minimizer won't let the maximizer reach ply 3; at ply 2, he'll move left so that the maximizer wins only if the card is a diamond. That means that the maximizer wins at the root just in case the card is either a club or a diamond.

A partial graph of the values for this game is as follows:

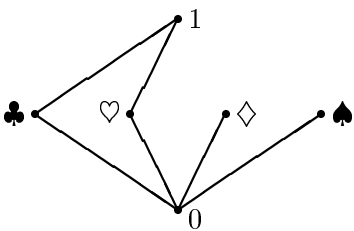

where we have included the crucial fact that $x \wedge y = 0$ if $x \neq y$ (since the minimizer knows the card) and $\heartsuit \vee \clubsuit = 1$ because the maximizer can invoke his special rule. Other least upper bounds are not shown in the diagram. The maximizing function $\vee$ moves up the figure; the minimizing function $\wedge$ moves down.

The deep prune fails because we can't "push" the value $\clubsuit \wedge 0$ past the $\heartsuit$ to get to the ♣ near the root. Somewhat more precisely, the problem is that

$$\heartsuit = \heartsuit \vee (\clubsuit \wedge 0) \neq (\heartsuit \wedge \clubsuit) \vee (\heartsuit \wedge 0) = 0$$

This suggests the following:

**Proposition 6.3.2 (Ginsberg & Jaffray, 2001)** *For a game $G$, the following conditions are equivalent:*

1. *$G$ allows $\alpha$-$\beta$ pruning for the minimizer.*

2. *$G$ allows $\alpha$-$\beta$ pruning for the maximizer.*

3. *$G$ allows pruning.*

4. *$(V, \wedge, \vee)$ is a distributive lattice.*

**Proof.** As before, we show only that the first and fourth conditions are equivalent. Since pruning implies shallow pruning (take $i = 0$ in the definition), it follows that the first condition implies that $(V, \wedge, \vee)$ is a lattice.

From deep pruning for the minimizer with $i = 1$, we have that if $x \leq y$, then for any $z_1, z_2, T$,

$$x \wedge (z_1 \vee (z_2 \wedge (y \vee T))) = x \wedge (z_1 \vee z_2)$$

Now take $y = T = x$ to get

$$x \wedge (z_1 \vee (z_2 \wedge x)) = x \wedge (z_1 \vee z_2) \tag{16}$$

It follows that each top level term in the left hand side of (16) is greater than or equal to the right hand side; specifically

$$z_1 \vee (z_2 \wedge x) \geq x \wedge (z_1 \vee z_2). \tag{17}$$

We claim that this implies that the lattice in question is distributive.

To see this, let $u, v, w \in V$. Now take $z_1 = u \wedge w$, $z_2 = v$ and $x = w$ in (17) to get

$$(u \wedge w) \vee (v \wedge w) \geq w \wedge ((u \wedge w) \vee v) \tag{18}$$

But $v \vee (u \wedge w) \geq w \wedge (v \vee u)$ is an instance of (17), and combining this with (18) gives us

$$
\begin{aligned}
(u \wedge w) \vee (v \wedge w) &\geq w \wedge ((u \wedge w) \vee v) \\
&\geq w \wedge w \wedge (v \vee u) \\
&= w \wedge (v \vee u)
\end{aligned}
$$

This is the hard direction; $w \wedge (v \vee u) \geq (u \wedge w) \vee (v \wedge w)$ for any lattice because $w \wedge (v \vee u) \geq u \wedge w$ and $w \wedge (v \vee u) \geq v \wedge w$ individually. Thus $w \wedge (v \vee u) = (u \wedge w) \vee (v \wedge w)$, and deep pruning implies that the lattice is distributive.

For the converse, if the lattice is distributive and $x \leq y$, then

$$
\begin{aligned}
x \wedge (z_1 \vee (z_2 \wedge (y \vee T))) &= (x \wedge z_1) \vee (x \wedge z_2 \wedge (y \vee T)) \\
&= (x \wedge z_1) \vee (x \wedge z_2) \\
&= x \wedge (z_1 \vee z_2)
\end{aligned}
$$

where the second equality is a consequence of the fact that $x \leq (y \vee T)$, so that $x = x \wedge (y \vee T)$. This validates pruning for $i = 1$; deeper cases are similar. $\quad\square$

Finally, note that in games where this result applies, we can continue to use Algorithms 2.2.5 or 2.3.3 without modification, since the prunes that they endorse continue to be sound as the game tree is expanded.

### 6.4 Application to imperfect information

In order to apply these ideas to games of imperfect information treated as in Section 5, we need to show that the value set introduced there is a (hopefully distributive) lattice.

To do this, recall that there is redundant information in an arbitrary element $\mathcal{F}$ of $2^{2^S}$, since if $\mathcal{F}$ contains both $T$ and $U$ with $T \subseteq U$ (in other words, the maximizer can play for either $T$ or for $U$ but $U$ is properly better), the set $T$ can be removed from $\mathcal{F}$ without affecting the maximizer's options in any interesting way. This suggests the following:

**Definition 6.4.1** *Let $\mathcal{F} \in 2^{2^S}$ for an arbitrary set $S$. We will say that $\mathcal{F}$ is reduced if there are no $T, U \in \mathcal{F}$ with $T \subseteq U$. We will say that $\mathcal{F}_1$ is a reduction of $\mathcal{F}_2$ if $\mathcal{F}_1$ is reduced and $\mathcal{F}_1 \subseteq \mathcal{F}_2$.*

**Lemma 6.4.2** *Every $\mathcal{F} \in 2^{2^S}$ has a unique reduction.*

**Proof.** This is immediate; just remove the subsumed elements from $\mathcal{F}$. $\qquad\square$

We will denote the reduction of $\mathcal{F}$ by $r(\mathcal{F})$.

Armed with this definition, we can now modify Definition 5.0.6 in the obvious way, replacing the value set $V$ with the set of reduced elements of $V$ and the maximizing and minimizing functions (8) and (9) with the reduced versions thereof, so that

$$\max(\mathcal{F}, \mathcal{G}) = r(\mathcal{F} \cup \mathcal{G}) \tag{19}$$

and

$$\min(\{F_i\}, \{G_j\}) = r(\{F_i \cap G_j\}) \tag{20}$$

Remember that we typically write $\vee$ for max and $\wedge$ for min.

**Proposition 6.4.3** *Given the above definitions, $(V, \vee, \wedge)$ is a distributive lattice.*

**Proof.** We need to show that max and min as defined above are commutative, associative, and idempotent, that they distribute with respect to one another, and that the absorption identity (12) is satisfied. Since the reduction operator clearly commutes with the initial definitions of max and min, commutativity, associativity and distributivity are obvious, as is the fact that $\vee$ is idempotent. To see that $\wedge$ is idempotent, we have

$$\mathcal{F} \wedge \mathcal{F} = r(\{\min(F_i, F_j)\}) = r(\{F_i \cap F_j\})$$

but each element of the set on the righthand side is a subset of $F_i \cap F_i$ so

$$\mathcal{F} \wedge \mathcal{F} = r(\{F_i\}) = r(\mathcal{F}) = \mathcal{F}.$$

For the absorption identity, we need to show that

$$\mathcal{F} \vee (\mathcal{F} \wedge \mathcal{G}) = \mathcal{F}$$

But

$$\mathcal{F} \wedge \mathcal{G} = r\{F_i \cap G_j\}$$

so

$$\begin{aligned}
\mathcal{F} \vee (\mathcal{F} \wedge \mathcal{G}) &= r(\mathcal{F} \vee r\{F_i \cap G_j\}) \\
&= r(\{F_i\} \cup \{F_i \cap G_j\}) \\
&= r(\{F_i\}) \\
&= r(\mathcal{F}) \\
&= \mathcal{F}
\end{aligned}$$

since, once again, each element of $\mathcal{F} \wedge \mathcal{G}$ is subsumed by the corresponding $F_i$. □

It follows that an implementation designed to compute the value of an imperfect information game as described by Theorem 5.0.7 can indeed use $\alpha$-$\beta$ pruning to speed the computation.

## 6.5 Bridge implementation

Given this body of theory, we implemented a single-dummy version of GIB's double-dummy search engine. Not surprisingly, the most difficult element of the implementation was building efficient data structures for the manipulation of elements of $2^{2^S}$.

To handle this, we represented each element of $S$ as a conjunction. We first identified one of the two hidden hands $H$, and then for each card $c$, would write $c$ if $c$ were held by $H$ and $\neg c$ if $c$ were not held by $H$. An element of $2^S$ was then taken to be a disjunctive combination of these conjunctions, and an element of $2^{2^S}$ was taken to be a list of such disjunctions. The advantage of this representation was that logical inference could be used to construct the reduction of any such list.

In order to make this inference as efficient as possible, the disjunctions themselves were represented as *binary decision diagrams*, or BDD's (Lind-Nielsen, 2000). There are a variety of public domain implementations of BDD's available, and we used one provided by Lind-Nielsen (Lind-Nielsen, 2000).[17]

The resulting implementation solves small endings (perhaps 16 cards left in total) quickly but for larger endings, the running times come to be dominated by the BDD computations; this is hardly surprising, since the size of individual BDDs can be exponential in the size of $S$ (the number of possible distributions of the unseen cards). We found that we were generally able to solve 32-card endings in about a minute, but that the running times were increasing by two orders of magnitude as each additional card was added.

This is both good news and bad news. Viewed positively, the performance of the system as constructed is far superior to the performance of preceding attempts to deal with the imperfect information arising in bridge. Frank et.al, for example, are only capable of solving single suit combinations (13 cards left, give or take) using an algorithm that appears to take several minutes to run (Frank, Basin, & Matsubara, 1998). They subsequently improve the performance to an average time of 0.6 seconds (Frank et al., 2000), but are still restricted to problems that are too small to be of much use to a program intended to play the complete game.

---

17. We tried a variety of non-BDD based implementations as well. The BDD-based implementation was far faster than any of the others.

That's the good news. The bad news is that a program capable only of solving an 8-card ending in a minute is inappropriate for production use. GIB is a production program, expected to play bridge at human speeds. Another approach was therefore needed.

## 7. Solving single-dummy problems in practice

### 7.1 Achievable sets

The key to practical application of the ideas in the previous section is the realization that when it comes time to make a play, a single element of $\mathcal{F}$ must be selected: if you can play for West to have the ♠Q or for each player to have three hearts but cannot cater to both possibilities simultaneously, you eventually have to actually make the choice.

**Definition 7.1.1** *Suppose that the value of the imperfect information game for $G$ is $\mathcal{F}$. Given a specific $A \subseteq S$, we will say that $A$ is* achievable *if there is some $F \in \mathcal{F}$ for which $A \subseteq F$.*

In other words, the set $A$ of situations is achievable if the maximizer has a plan that wins for all elements of $A$.

**Definition 7.1.2** *Given a set $S$ of situations, a* payoff *function for $S$ is any function $f : 2^S \to \mathbb{R}$ such that $f(U) \leq f(T)$ whenever $U \subseteq T$.*

The payoff function evaluates potential achievable sets.

**Definition 7.1.3** *Let $G$ be a game and $S$ the associated set of situations. If $f$ is a payoff function for $S$, a* solution *to $G$ under $f$ is any achievable set $A$ for which $f(A)$ is maximal.*

In practice, we need not find the actual value of the game; finding a solution to $G$ under an appropriate payoff function suffices. In bridge, the payoff function is presumably the probability that the cards are dealt as in the set $A$; this function clearly increases with increasing set size as required by Definition 7.1.2 and can be evaluated in practice using the Monte Carlo sample of Section 3.

Instead of finding the solution to an imperfect information game, suppose instead that we have a Monte Carlo sample for the game consisting of a set of situations $S = \{s_i\}$ that is ordered as $i = 0, \ldots, n$. We can now produce an achievable set $A$ as follows:

**Algorithm 7.1.4** *To construct a maximal achievable set $A$ from a sequence $\langle s_0, \ldots, s_n \rangle$ of situations:*

1. *Set $A = \varnothing$.*
2. *For $i = 0, \ldots, n$, if $A \cup \{s_i\}$ is achievable, set $A = A \cup \{s_i\}$.*

The algorithm constructs the achievable set in a greedy fashion, gradually adding elements of $S$ to $A$ until no more can be added.

**Definition 7.1.5** *Given a game $G$ and a sequence $S$ of situations, the* achievable set *induced by $S$ for $G$ is the set constructed by Algorithm 7.1.4.*

From a computational point of view, the expensive step in the algorithm is determining whether or not the set $A \cup \{s_i\}$ is achievable. This is relatively straightforward, however, since the focus on a specific set effectively replaces the game $G$ with a new game with values in $\{0, 1\}$. At any particular node $n$, if expanding $n$ demonstrates that $A \cup \{s_i\}$ is not achievable, the value of the game is zero. If expanding $n$ indicates that $A \cup \{s_i\}$ is achievable once $n$ is reached, then the value of the node $n$ is one. Although the search space is unchanged from that of the original imperfect information game as in Definition 5.0.6, there is no longer any need to manipulate complex values, and the check for achievability is therefore tractable in practice.

Let me illustrate this by returning to our usual example of Section 5. Here is the fully evaluated tree once again:

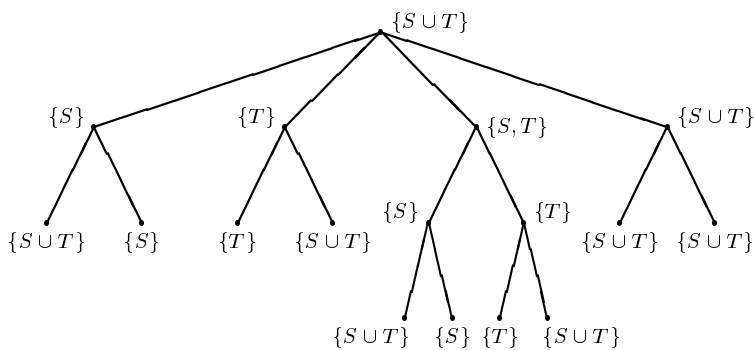

Note that we have replaced the value at the root with its reduction.

Now suppose that we view the set of positions as containing only two elements, $s \in S$ and $t \in T$. Presumably West holds the ♠Q in $s$, and East holds three hearts in $t$. If the ordering chosen is $\langle s, t \rangle$, then we first try to achieve $\{s\}$. In this context, a node $n$ is a win for the maximizer if either the maximizer can indeed win at $n$ or $s$ is no longer possible (in which case the maximizer's ability to achieve $\{s\}$ is undiminished). The game tree becomes:

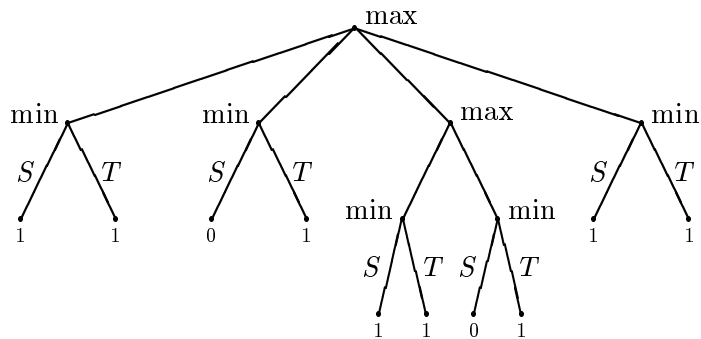

All of the $T$ branches are wins for the maximizer (who is concerned with $s$ only), and the $S$ branches are wins just in case the maximizer does indeed win (as he does if he guesses right at either of the first two plies). Backing up the values gives us:

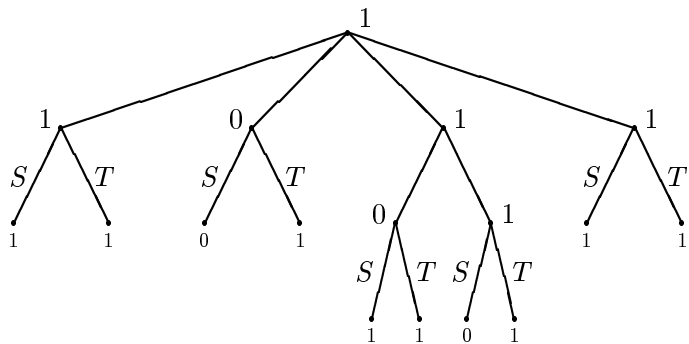

This indicates (correctly) that the maximizer can achieve $s$ provided that he doesn't decide to play for $T$ at the root of the tree. Note that this analysis is a straight minimax, allowing fast algorithms to be applied while avoiding the manipulation of elements of $2^{2^S}$ described in the previous section.

Now we add $t$ to our achievable set, which thus becomes $\{s, t\}$. The maximizer wins only if he really does win (and not just because he isn't interested in $T$ any more), and the basic tree becomes:

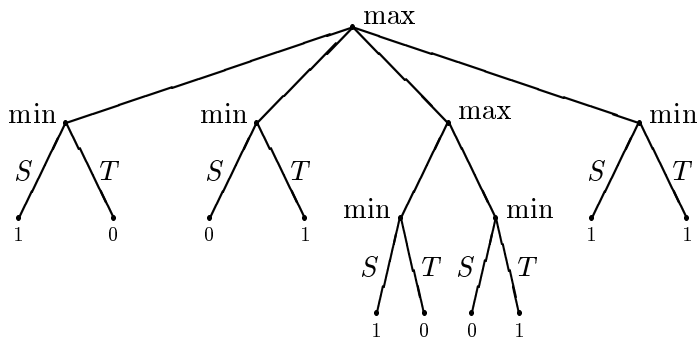

Backing up the values gives:

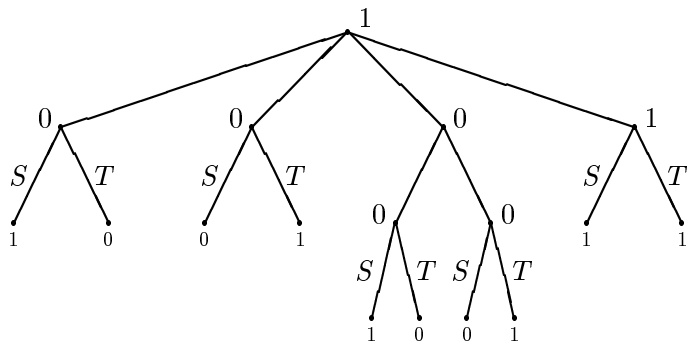

The maximizer can achieve the extended result only by making the rightmost move, as desired.

What if the rightmost branch did not exist, so that the maximizer were unable to combine his chances? Now the value of the root node in the above tree is 0, so that $\{s, t\}$ is not achievable. The maximal achievable set returned by the algorithm would be $S$; had the ordering been $\langle t, s \rangle$ instead, an alternative maximal achievable set of $T$ would have been returned instead.

In any event, we have:

**Proposition 7.1.6** *Given a game $G$ and a sequence $S$ of situations, let $A$ be the achievable set induced by $S$ for $G$. Then no proper superset of $A$ in $S$ is achievable.*

**Proof.** This is straightforward. For any element $s \in S - A$, we know that $U \cup \{s\}$ is not achievable for some $U \subseteq A$. Thus $A \cup \{s\}$ is not achievable as well. □

Algorithm 7.1.4 allows us to construct maximal achievable sets relative to our Monte Carlo sample; recall that we are taking our sequence $S$ of situations to be any ordering of the sample itself. In practice, however, it is important not to focus too sharply on the sample itself, lest the eventual achievable set constructed overfit irrelevant probabilistic characteristics of that sample. This can be accomplished by replacing the simple union in step 2 of the algorithm with some more complicated operation that captures the idea of "situations that are either like $s_i$ or like those already in $A$." In bridge, for example, $A$ might be all situations where West has two or three hearts, and $s_i$ might be some new situation where West has four hearts. The generalized union would be situations where West has two, three or four hearts. If this more general set is not achievable, another attempt could be made with the simple union. If we denote the "general union" by $\oplus$, Algorithm 7.1.4 becomes:

**Algorithm 7.1.7** *To construct an achievable set $A$ from a sequence $\langle s_0, \ldots, s_n \rangle$ of situations:*

1. *Set $A = \varnothing$.*
2. *For $i = 0, \ldots, n$:*
   - (a) *If $A \oplus \{s_i\}$ is achievable, set $A = A \oplus \{s_i\}$.*
   - (b) *Otherwise, if $A \cup \{s_i\}$ is achievable, set $A = A \cup \{s_i\}$.*

This algorithm can be used in practice to find achievable sets that are either maximal or effectively so over the set of all possible instances, not just those appearing in the Monte Carlo sample.

## 7.2 Maximizing the payoff

It remains to find not just maximal achievable sets, but ones that approximate the solution to the game in question given a particular payoff function.

To understand how we do this, let me draw an analogy between the problem we are trying to solve and resource-constrained project scheduling (RCPS). In RCPS, one has a list of tasks to be performed, together with ordering constraints saying that certain tasks need to be performed before others. In addition, each task uses a certain quantity of various resources; there are limitations on the availability of any particular resource at any particular time. As an example, building an aircraft wing may involve fabricating the top and bottom flight surfaces, building the aileron, and attaching the two. It should be clear that the aileron cannot be attached until both it and the wing have been constructed. Building each section may involve the use of three sheetmetal workers, but only five may be available in general.

The goal in an RCPS problem is typically to minimize the length of the schedule (often called the *makespan*) without exceeding the resource limits. In building a wing, it is more efficient (and more cost effective) to build it quickly than slowly.

Many production scheduling systems try to minimize makespan by building the schedule from the initial time forward. At each point, they select a task all of whose predecessors have been scheduled, and then schedule that task as early as possible given the previously scheduled tasks and the resource constraints. Scheduling the tasks in this way produces a locally optimal schedule that may be improved by modifying the order in which the tasks are selected for scheduling.

One method for finding an appropriate modification to the selection order is known as *squeaky wheel optimization*, or SWO (Joslin & Clements, 1999). In SWO, a locally optimal schedule is examined to determine which tasks are scheduled most suboptimally relative to some overall metric; those tasks are deemed to "squeak" and are then advanced in the task list so that they are scheduled earlier when the schedule is reconstructed. This process is repeated, producing a variety of candidate solutions to the scheduling problem at hand; one of these schedules is typically optimal or nearly so.

Applying SWO to our game-playing problem is relatively straightforward.[18] When we use Algorithm 7.1.7 to construct an achievable set, we also construct as a byproduct a list of sample elements to which that achievable set cannot be extended; moving elements of this list forward in the sequence of $\langle s_0, \ldots, s_n \rangle$ will cause them to be more likely to be included in the achievable set $A$ if the algorithm is reinvoked. The weights assigned to the failing sequence elements can be constructed by determining how representative each particular element is of the remainder of the sample.

Returning to our example, suppose that the set $S$ (where West has the ♠Q) has a single representative $s_1$ in the Monte Carlo sample (presumably this means it is unlikely for West to hold the card in question), while $T$ has five such representatives $t_1$, $t_2$, $t_3$, $t_4$ and $t_5$. Suppose also that the initial ordering of the six elements is $\langle s_1, t_4, t_2, t_1, t_5, t_3 \rangle$.

Assuming that the maximizer loses his rightmost option (so that he cannot cater to $S$ and $T$ simultaneously), the maximal achievable set corresponding to this ordering is $S$. An examination now reveals that all of the $t_i$'s could have been achieved but weren't; in SWO terms, these elements of the sample "squeak."

At the next iteration, the priorities of the $t_i$'s are increased by moving them forward in the sequence, while the priority of $s_1$ falls. Perhaps the new ordering is $\langle t_4, t_2, s_1, t_1, t_5, t_3 \rangle$.

This ordering can be easily seen to lead to the maximal achievable set $T$; $S \cup T$ is still unachievable. But the payoff assigned to $T$ is likely to be much better than that assigned to $S$ (a probability of 0.8 instead of 0.2, if the Monte Carlo sample itself is unweighted). It is in this way that SWO allows us to find a globally optimal (or nearly so) achievable set.

---

18. Squeaky wheel optimization was developed at the University of Oregon; the patent application for the technique has been allowed by the U.S. Patent and Trademark Office. The University's interests in SWO are licensed exclusively to On Time Systems, Inc. for use in scheduling and related applications, and to Just Write, Inc. for use in bridge-playing systems.

## 7.3 Results

Our implementation of GIB's cardplay when declarer is based on the ideas described above. (As a defender, a direct Monte Carlo approach appears preferable because enough information is typically available about declarer's hand to make the double-dummy assumption reasonably valid.) The implementation is fast enough to conform to the time requirements placed on a production program (roughly one CPU minute to play each deal).

Evaluating the impact of these ideas on GIB's cardplay is difficult, since declarer play is already the strongest aspect of its game. In extended matches between the two versions of GIB, the approach based on the ideas described here beats the Monte-Carlo based version by approximately 0.1 IMPs/deal, but there is a great deal of noise in the data because most of the swings correspond to differences in bidding or defensive play. It is possible to remove some of these differences artificially (requiring the bidding to be identical both times the deal is played, for example), but defensive differences remain. Nevertheless, GIB is currently a strong enough player that the 0.1 IMPs/deal difference is significant.

The situation on problem deals, such as those from the par contests or from the Gitelman sets, is much clearer. In addition, many of the deals that GIB gets "wrong" are in fact deals that GIB plays correctly but that the problem composers play incorrectly (Gitelman or, in the case of the par contests, Swiss bridge expert Pietro Bernasconi). In the following table, we have been generous with all parties, deeming a line to be correct if it is not clearly inferior to another. Let me point out that the designers of the problems are *attempting* to construct deals where there is a unique solution (the "answer" to the test they are posing the solver), so that a deal with multiple solutions is in fact one that the composer has already misanalyzed.

| Source | size | BB | GIB$_{MC}$ | GIB$_{SWO}$ | composer | ambiguous |
|---|---|---|---|---|---|---|
| BM level 1 | 36 | 16 | 31 | 36 | 35 | 0 |
| level 2 | 36 | 8 | 23 | 34 | 34 | 1 |
| level 3 | 36 | 2 | 12 | 34 | 34 | 2 |
| level 4 | 36 | 1 | 21 | 31 | 34 | 4 |
| level 5 | 36 | 4 | 13 | 28 | 34 | 5 |
| 1998 par contest | 12 | 0 | 5 | 11 | 12 | 2 |
| 1990 par contest | 18 | 0 | 8 | 14 | 17 | 3 |

The rows are in order of increasing difficulty; it was universally felt among the human competitors that the deals in the 1990 par contest were far more difficult than those in 1998. The columns are as follows:

**Source**      is the source from which the problems were obtained.
**Size**        is the number of problems available from this particular source.
**BB**          gives the number of problems solved correctly by Bridge Baron 6.
GIB$_{MC}$      gives the number solved correctly by GIB using a Monte Carlo approach.
GIB$_{SWO}$     gives the number solved correctly by GIB using SWO and achievable sets.
**composer**    gives the number solved correctly by the composer (in that the intended solution was the best one available).
**ambiguous**   gives the number misanalyzed by the composer (in that multiple solutions exist).

Note, incidentally, that GIB's performance is still less than perfect on these problems. The reason is that GIB's sample may be skewed in some way, or that SWO may fail to find a global optimum among the set of possible achievable sets.

## 8. Conclusion

### 8.1 GIB compared

**Other programs**   GIB participated in both the 1998 and the 2000 World Computer Bridge Championships. (There was no 1999 event.) Play was organized with each machine playing two hands and the competitors being trusted not to cheat by "peeking" at partner's cards or those of the opponents.[19]

Each tournament began with a complete round robin among the programs, with the top four programs continuing to a knockout phase. The matches in the round robin were quite short, and it was expected that bridge's stochastic element would keep any program from being completely dominant.

While this may have been true in theory, in practice GIB dominated both round robins, winning all of its matches in 1998 and all but one in 2000. The round robin results from the 2000 event were as follows:[20]

| | GIB | WB | MICRO | BUFF | Q-PLUS | CHIP | BARON | M'LARK | Total |
|---|---|---|---|---|---|---|---|---|---|
| GIB | – | 14 | 11 | 16 | 7 | 19 | 16 | 17 | 100 |
| WBRIDGE | 6 | – | 19 | 13 | 16 | 7 | 18 | 20 | 99 |
| MICRO | 9 | 1 | – | 18 | 15 | 15 | 13 | 20 | 91 |
| BUFF | 4 | 7 | 2 | – | 12 | 20 | 5 | 20 | 70 |
| Q-PLUS | 13 | 4 | 5 | 8 | – | 11 | 14 | 11 | 66 |
| BLUE CHIP | 1 | 13 | 5 | 0 | 9 | – | 11 | 20 | 59 |
| BARON | 4 | 2 | 7 | 15 | 6 | 9 | – | 14 | 57 |
| MEADOWLARK | 3 | 0 | 0 | 0 | 9 | 0 | 6 | – | 18 |

Each match was converted first to IMPs and then to *victory points*, or VPs, with the two competing programs sharing the 20 VPs available in each match. The first entry in the above table indicates that GIB beat WBRIDGE by 14 VPs to 6; the fourth that GIB lost to Q-PLUS BRIDGE by 7 VPs to 13. (This is GIB's only loss ever to another program in tournament play.)

In the 1998 knockout phase, GIB beat Bridge Baron in the semifinals by 84 IMPs over 48 deals. Had the programs been evenly matched, the IMP difference could be expected to be normally distributed, and the observed 84 IMP difference would be a 2.2 standard deviation event. GIB then beat Q-Plus Bridge in the finals by 63 IMPs over 64 deals (a 1.4 standard deviation event). In 2000, it beat Bridge Buff by 39 IMPs over 48 deals in the semifinals (a 1.0 standard deviation event) and then beat WBRIDGE by 101 IMPs over 58 deals (a 2.6 standard deviation event). The finals had been scheduled to run 64 deals, but WBRIDGE conceded after 58 had been played.

---

19. Starting with the 2001 event, each computer will handle only one of the four players, although there is still no attempt to prevent the (networked) computers from transmitting illegal information between partners.

20. There were eight competitors in the event:   GIB (www.gibware.com), Hans Leber's Q-PLUS (www.q-plus.com), Tomio and Yumiko Uchida's MICRO BRIDGE (www.threeweb.ad.jp/~mcbridge), Mike Whittaker and Ian Trackman's BLUE CHIP bridge (www.bluechipbridge.co.uk), Rod Ludwig's MEADOWLARK bridge (rrnet.com/meadowlark), bridge BARON (www.bridgebaron.com), and two newcomers: Doug Bannion's bridge BUFF (www.bridgebuff.com) and Yves Costel's WBRIDGE (ourworld.compuserve.com/homepages/yvescostel).

The most publicized deal from the final was this one, an extremely difficult deal that both programs played moderately well. GIB reached a better contract and was aided somewhat by WBRIDGE's misdefence in a moderately complex situation.

```
                    ♠ K Q 9
                    ♡ A Q J
                    ◇ 9 6 4 3 2
                    ♣ 8 6
    ♠ 10 6                          ♠ 8 7 3 2
    ♡ 10 9 2                        ♡ 7 5 3
    ◇ 10                            ◇ A K Q J 8 5
    ♣ A J 10 9 5 3 2                ♣ —
                    ♠ A J 5 4
                    ♡ K 8 6 4
                    ◇ 7
                    ♣ K Q 7 4
```

When WBRIDGE played the North-South cards and GIB was East-West, North opened 1◇ and eventually played in three notrump, committing to taking nine tricks. The GIB East started with four rounds of diamonds as South discarded two clubs and . . . ?

Looking at all four hands, the contract is cold; South can discard another club and East has none to play. There are thus nine tricks: four in each of hearts and spades, and the diamond nine.

Give East a club, however, and the contract rates to be down no less than four since the defense will be able to take at least four club tricks. WBRIDGE decided to play safe, keeping the ♣KQ and discarding a heart. There are now only eight tricks and the contract was down one.

The bidding and play were more interesting when GIB was N-S. North opened 1NT, showing 11–14 HCP without four hearts or spades unless exactly three cards were held in every other suit. East overcalled a natural 2◇ and South cue bid 3◇, showing weakness in diamonds and asking North to bid a 4-card heart or spade suit if he had one.

North has no good bid at this point. Bidding 3NT with five small diamonds rates to be wrong and 4♣ is clearly out of the question. GIB's simulation suggested that 3♠ (ostensibly showing four of them) was the least of evils. South raised to 4♠, and East doubled, ending the auction.

East led a top diamond, and shifted to the ♡3, won by North's ♡Q. GIB now cashed the ♡J and led the ♣6, which East chose (wrongly) to ruff. WBRIDGE now led the ◇K as East, which was ruffed with the ♠J. GIB was now able to cash the ♠AK to produce:

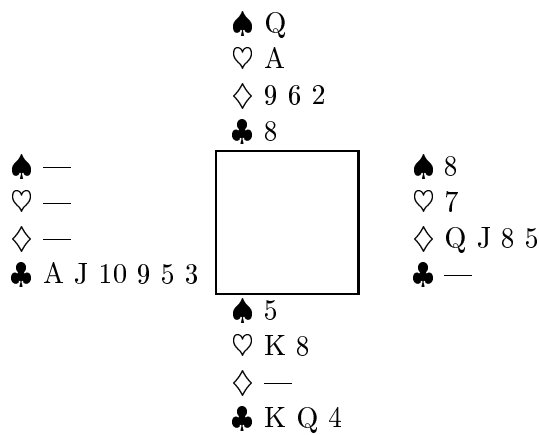

Knowing the position exactly, GIB needed five more tricks with North to lead. It ruffed a diamond, returned to the ♡A and drew East's trump with the ♠Q. Now a club forced an entry to the South hand, where the ♡K provided the tenth trick.

**Humans** GIB played a 14-deal demonstration match against human world champions Zia Mahmood and Michael Rosenberg[21] in the AAAI Hall of Champions in 1998, losing by a total of 6.4 IMPs (a 0.3 standard deviation event). Early versions of GIB also played on OKBridge, an internet bridge club with some 15,000 members.[22] After playing thousands of deals against human opponents of various levels, GIB's ranking was comparable to the OKBridge average.

It is probable that neither of these results is an accurate reflection of GIB's current strength. The Mahmood-Rosenberg match was extremely short and GIB appeared to have the best of the luck. The OKBridge interface has changed and the GIB 'OKbots' no longer function. The performance figures there are thus somewhat outdated, predating various recent improvements including all of the ideas in Sections 5–7. More interesting information will become available starting in late July of 2001, when GIB, paired with Gitelman and his regular partner Brad Moss, will begin a series of 64-deal matches against human opponents of varying skill levels.

## 8.2 Current and future work

Recent work on GIB has focused on its weakest areas: defensive cardplay and bidding. The bidding work has been and continues to be primarily a matter of extending the existing bidding database, although GIB's bidding language is also being changed from Standard American (a fairly natural system) to a variant of an artificial system called Moscito developed in Australia.[23] Moscito has very sharply defined meanings, making it ideal for use by a computer program, and is an "action" system, working hard to make the opponents' bidding as difficult as possible.

---

21. Mahmood and Rosenberg have won, among other titles, the 1995 Cap Volmac World Top Invitational Tournament. As remarked earlier, Rosenberg would also go on after the GIB match to win the Par Competition in which GIB finished 12th.

22. http://www.okbridge.com

23. GIB's version of Moscito is called *Moscito Byte*.

With regard to defensive cardplay, the key elements of high level defense are to make it hard for partner to make a mistake while making it easy for declarer to do so. Providing GIB with these abilities will involve an extra level of recursion in the cardplay, as each element of the Monte Carlo sample must now be considered from other players' points of view, as they generate and then analyze their own samples. These ideas have been implemented but currently lead to small performance degradations (approximately 0.05 IMPs/deal) because the computational cost of the recursive analyses require reducing the size of the Monte Carlo sample substantially. As processor speeds increase, it is reasonable to expect these ideas to bear significant fruit.

In 1997, Martel, a computer scientist himself, suggested that he expected GIB to be the best bridge player in the world in approximately 2003.[†] The work appears to be roughly on schedule.

## 8.3 Other games

I have left essentially untouched the question of to what extent the basic techniques we have discussed could be applied to games of imperfect information other than bridge.

The ideas that we have presented are likely to be the most applicable in games where the perfect information variant is tractable but computationally challenging, and the assumption that one's opponents are playing with perfect information is a reasonable one. This suggests that games like hearts and other trick-taking games will be amenable to our techniques, while games like poker (where it is essential to realize and exploit the fact that the opponents also have imperfect information) are likely to need other approaches.

## Acknowledgments

A great many people have contributed to the GIB project over the years. In the technical community, I would like to thank Jonathan Schaeffer, Rich Korf, David Etherington, Bart Massey and the other members of CIRL. In the bridge community, I have received invaluable assistance from Chip Martel, Rod Ludwig, Zia Mahmood, Andrew Robson, Alan Jaffray, Hans Kuijf, Fred Gitelman, Bob Hamman, Eric Rodwell, Jeff Goldsmith, Thomas Andrews and the members of the rec.games.bridge community. The work itself has been supported by Just Write, Inc., by DARPA/Rome Labs under contracts F30602-95-1-0023 and F30602-97-1-0294, and by the Boeing Company under contract AHQ569. To everyone who has contributed, whether named above or not, I owe my deepest appreciation.

## Appendix A. A summary of the rules of bridge

We give here a very brief summary of the rules of bridge. Readers wanting a more complete description are referred to any of the many excellent texts available (Sheinwold, 1996).

Bridge is a card game for four players, who are split into two pairs. Members of a single pair sit opposite one another, so that North-South form one pair and East-West the other.

The deck is distributed evenly among the players, so that each deal involves giving each player a *hand* of 13 cards. The game then proceeds through a bidding and a playing phase.

The playing phase consists of 13 *tricks*, with each player contributing one card to each trick in a clockwise fashion. The player who plays first to any trick is said to *lead* to that trick. The highest card of the suit led wins the trick (Ace is high and deuce low), unless a trump is played, in which case the highest trump wins the trick. The person who leads to a trick is free to lead any card he wishes; subsequent players must play a card of the suit led if they have one, and can play any card they choose if they don't. The winner of one trick leads to the next; the person who leads to the first trick (the *opening leader*) is determined during the bidding phase of the game.

The object of the card play phase is always for your partnership to take as many tricks as possible; there is no advantage to one partner's taking a trick over another, and the order in which the tricks are taken is irrelevant. After the opening leader plays the first card to the first trick, the player to his left places his cards face up on the table so that all of the other players can see them. This player is called the *dummy*, and when it is dummy's turn to play, dummy's partner (who can see the partnership's combined assets) selects the card to be played. Dummy's partner is called the *declarer* and the members of the other pair are called the *defenders*.

The purpose of the bidding phase is to identify trumps and the declarer, and also the *contract*, which will be described shortly. The opening leader is identified as well, and is the player to the declarer's left.

During the bidding phase, various contracts are proposed. The dealer has the first opportunity to propose a contract and subsequent opportunities are given to each player in a clockwise direction. Each player has many opportunities to suggest a contract during this phase of the game, which is called the *auction*. Each partnership is required to explain the meanings of their actions during the auction to the other side, if requested.

Each contract suggests a particular trump suit (or perhaps that there not be a trump suit at all). Each player suggesting a contract is committing his side to winning some particular number of the 13 available tricks. The minimum commitment is 7 tricks, so there are 35 possible contracts (each of 4 possible trumps, or no trumps, and seven possible commitments, from seven to thirteen tricks). These 35 contracts are ordered, which guarantees that the bidding phase will eventually terminate.

After the bidding phase is complete, the side that suggested the final contract is the *declaring side*. Of the two members of the declaring side, the one who first suggested the eventual trump suit (or no trumps) is the declarer. Play begins with the player to the declarer's left leading to the first trick.

After the hand is complete, there are two possible outcomes. If the declaring side took at least as many tricks as it committed to taking, the declaring side receives a positive score and the defending side an equal but negative score. There are substantial bonuses awarded for committing to taking particular numbers of tricks; in general, the larger the commitment, the larger the bonus. There are small bonuses awarded for winning tricks above and beyond the commitment.

If the declaring side failed to honor its commitment, it receives a negative score and the defenders receive an equal but positive score. The overall score in this case (where the declarer "goes down") is generally smaller than the overall score in the case where declarer "makes it" (i.e., honors his commitment).

## Appendix B. A new ending discovered by GIB

This deal occurred during a short IMP match between GIB and Bridge Baron.

```
                    ♠ 9 6
                    ♡ Q J 8 5
                    ♢ A Q 3
                    ♣ K J 10 8
  ♠ K Q J 8 7 5   ┌─────────────┐   ♠ 4 3
  ♡ 9 4 3         │             │   ♡ A 7 2
  ♢ 7             │             │   ♢ J 10 6 2
  ♣ 6 4 2         └─────────────┘   ♣ A Q 7 3
                    ♠ A 10 2
                    ♡ K 10 6
                    ♢ K 9 8 5 4
                    ♣ 9 5
```

With South (GIB) dealing at unfavorable vulnerability, the auction went P–2♠–X–P–3NT– all pass. (P is pass and X is double.) The opening lead was the ♠K, ducked by GIB, and Bridge Baron now switched to a small heart. East won the ace and returned to spades, GIB winning.

GIB cashed all the hearts, pitching a small club from its hand. It then tested the diamonds, learning of the bad break and winning the third diamond in hand. It then led the ♢9 in the following position:

```
                    ♠ —
                    ♡ —
                    ♢ —
                    ♣ K J 10 8
  ♠ Q             ┌─────────────┐   ♠ —
  ♡ —             │             │   ♡ —
  ♢ —             │             │   ♢ J
  ♣ ? ? ?         └─────────────┘   ♣ A ? ?
                    ♠ 10
                    ♡ —
                    ♢ 9 8
                    ♣ 9
```

When GIB pitched the *ten* of clubs from dummy (it had been aiming for this ending all along), the defenders were helpless to take more than two tricks independent of the location of the club queen. At the other table, Bridge Baron let GIB play in 2♠ making exactly, and GIB picked up 12 IMPs.

# References

Adelson-Velskiy, G., Arlazarov, V., & Donskoy, M. (1975). Some methods of controlling the tree search in chess programs. *Artificial Intelligence*, *6*, 361–371.

Bayardo, R. J., & Miranker, D. P. (1996). A complexity analysis of space-bounded learning algorithms for the constraint satisfaction problem. In *Proceedings of the Thirteenth National Conference on Artificial Intelligence*, pp. 298–304.

Billings, D., Papp, D., Schaeffer, J., & Szafron, D. (1998). Opponent modeling in poker. In *Proceedings of the Fifteenth National Conference on Artificial Intelligence*, pp. 493–499.

Blackwood, E. (1979). *Play of the Hand with Blackwood*. Bobbs-Merrill.

Eskes, O. (1997). GIB: Sensational breakthrough in bridge software. *IMP*, *8*(2).

Frank, I. (1998). *Search and Planning under Incomplete Information: A Study Using Bridge Card Play*. Springer-Verlag, Berlin.

Frank, I., & Basin, D. (1998). Search in games with incomplete information: A case study using bridge card play. *Artificial Intelligence*, *100*, 87–123.

Frank, I., Basin, D., & Bundy, A. (2000). Combining knowledge and search to solve single-suit bridge. In *Proceedings of the Sixteenth National Conference on Artificial Intelligence*, pp. 195–200.

Frank, I., Basin, D., & Matsubara, H. (1998). Finding optimal strategies for imperfect information games. In *Proceedings of the Fifteenth National Conference on Artificial Intelligence*, pp. 500–507.

Ginsberg, M. L. (1993). Dynamic backtracking. *Journal of Artificial Intelligence Research*, *1*, 25–46.

Ginsberg, M. L., & Harvey, W. D. (1992). Iterative broadening. *Artificial Intelligence*, *55*, 367–383.

Ginsberg, M. L., & Jaffray, A. (2001). Alpha-beta pruning under partial orders. In *Games of No Chance II*. To appear.

Grätzer, G. (1978). *General Lattice Theory*. Birkhäuser Verlag, Basel.

Greenblatt, R., Eastlake, D., & Crocker, S. (1967). The greenblatt chess program. In *Fall Joint Computer Conference 31*, pp. 801–810.

Joslin, D. E., & Clements, D. P. (1999). Squeaky wheel optimization. *Journal of Artificial Intelligence Research*, *10*, 353–373.

Koller, D., & Pfeffer, A. (1995). Generating and solving imperfect information games. In *Proceedings of the Fourteenth International Joint Conference on Artificial Intelligence*, pp. 1185–1192.

Levy, D. N. (1989). The million pound bridge program. In Levy, D., & Beal, D. (Eds.), *Heuristic Programming in Artificial Intelligence*, Asilomar, CA. Ellis Horwood.

Lind-Nielsen, J. (2000). BuDDy: Binary Decision Diagram package. Tech. rep., Department of Information Technology, Technical University of Denmark, DK-2800 Lyngby, Denmark.

Lindelöf, T. (1983). *COBRA: The Computer-Designed Bidding System*. Gollancz, London.

Marsland, T. A. (1986). A review of game-tree pruning. *J. Intl. Computer Chess Assn.*, *9*(1), 3–19.

McAllester, D. A. (1988). Conspiracy numbers for min-max searching. *Artificial Intelligence*, *35*, 287–310.

Pearl, J. (1980). Asymptotic properties of minimax trees and game-searching procedures. *Artificial Intelligence*, *14*(2), 113–138.

Pearl, J. (1982). A solution for the branching factor of the alpha-beta pruning algorithm and its optimality. *Comm. ACM*, *25*(8), 559–564.

Plaat, A., Schaeffer, J., Pijls, W., & de Bruin, A. (1996). Exploiting graph properties of game trees. In *Proceedings of the Thirteenth National Conference on Artificial Intelligence*, pp. 234–239.

Schaeffer, J. (1997). *One Jump Ahead: Challenging Human Supremacy in Checkers*. Springer-Verlag, New York.

Sheinwold, A. (1996). *Five Weeks to Winning Bridge*. Pocket Books.

Smith, S. J., Nau, D. S., & Throop, T. (1996). Total-order multi-agent task-network planning for contract bridge. In *Proceedings of the Thirteenth National Conference on Artificial Intelligence*, Stanford, California.

Stallman, R. M., & Sussman, G. J. (1977). Forward reasoning and dependency-directed backtracking in a system for computer-aided circuit analysis. *Artificial Intelligence*, *9*, 135–196.

Sterling, L., & Nygate, Y. (1990). PYTHON: An expert squeezer. *J. Logic Programming*, *8*, 21–40.

Wilkins, D. E. (1980). Using patterns and plans in chess. *Artificial Intelligence*, *14*, 165–203.